\documentclass[final,12pt]{msml2021} 

\usepackage{graphicx}
\usepackage{caption} 
\usepackage{hyperref}
\usepackage{url}
\usepackage{color}

\usepackage{chngcntr}

\def\I{\mathbf{I}}
\def\J{\mathbf{J}}
\def\K{\mathbf{K}}

\def\N{\mathbf{N}}

\def\U{\mathbf{U}}
\def\V{\mathbf{V}}
\def\W{\mathbf{W}}

\def\a{\mathbf{a}}
\def\b{\mathbf{b}}

\def\e{\mathbf{e}}
\def\f{\mathbf{f}}
\def\g{\mathbf{g}}

\def\k{\mathbf{k}}

\def\v{\mathbf{v}}
\def\w{\mathbf{w}}
\def\x{\mathbf{x}}
\def\y{\mathbf{y}}
\def\z{\mathbf{z}}

\def\bTheta{\boldsymbol{\Theta}}

\def\bSigma{\boldsymbol{\Sigma}}

\def\bgamma{\boldsymbol{\gamma}}
\def\bdelta{\boldsymbol{\delta}}

\def\btheta{\boldsymbol{\theta}}

\def\supr#1{\underset{#1}{\textrm{sup}}}

\def\Exp{\mathbb{E}}
\def\Expp#1{\Exp\left[#1\right]}
\newcommand{\Exppp}[2]{\Exp_{#1}\left[#2\right]}

\def\CRLB_pe{\text{CRLB}}
\def\CRLB_pe{\text{CRLB}(\mathbf{p}_{e})}

\def\N02{\frac{N_0}{2}}

\def\0{\mathbf{0}}
\def\1{\mathbf{1}}

\def\0{\mathbf{0}}
\def\1{\mathbf{1}}

\newcommand{\tomt}[1]{\textcolor{black}{#1}}

\newcommand{\tomtb}[1]{\textcolor{black}{#1}}
\newcommand{\tomtr}[1]{\textcolor{black}{#1}}

\title[Kernel-Based Smoothness Analysis of Residual Networks]{Kernel-Based Smoothness Analysis of Residual Networks}
\usepackage{times}

\msmlauthor{%
 \Name{Tom Tirer} \Email{tomtirer@mail.tau.ac.il}\\
 \addr Tel Aviv University
 \AND
 \Name{Joan Bruna} \Email{bruna@cims.nyu.edu}\\
 \addr New York University%
 \AND
 \Name{Raja Giryes} \Email{raja@tauex.tau.ac.il}\\
 \addr Tel Aviv University%
}

\makeatletter
 \let\Ginclude@graphics\@org@Ginclude@graphics
\makeatother

\begin{document}

\maketitle

\begin{abstract}%
A major factor in the success of deep neural networks is the use of sophisticated architectures rather than the classical multilayer perceptron (MLP). Residual networks (ResNets) stand out among these powerful modern architectures.
Previous works focused on the optimization advantages of deep ResNets over deep MLPs. 
In this paper, we show another distinction between the two models, namely, a 
tendency of ResNets to promote smoother interpolations than MLPs. 
We analyze this phenomenon via the neural tangent kernel (NTK) approach. 
First, we compute the NTK for a considered ResNet model and prove its stability during gradient descent training. Then, we show by various evaluation methodologies that 
\tomtb{for ReLU activations} 
the NTK of ResNet, and its kernel regression results, are smoother than the ones of MLP. 
The better smoothness observed in our analysis may explain the better generalization ability of ResNets 
and the practice of moderately attenuating the residual blocks.

\end{abstract}


\section{Introduction}
\label{sec:int}

Deep neural networks have led to a major improvement in various fields. The advance in the network performance is tightly related to the introduction of various novel architectures \citep{Krizhevsky12ImageNet,Simonyan15VGG,he2016deep,Huang17Densely,EfficientNet19Tan}. A prominent model among them, which has led to a major leap in performance, is the deep residual network, known also as ResNet \citep{he2016deep}. It has introduced the usage of the skip connection, i.e., an identity path in the network that adds to the output features of a given layer its input features. This simple change enables effectively training much deeper networks, which eventually leads to improved results.

Different efforts were dedicated to explaining the success of ResNets. These mainly focused on the optimization aspect of ResNets, namely, e.g., claiming that it is ``easier'' to train a network with skip-connections as it enjoys a better loss surface \citep{Li18Visualizing} or that ResNets overcome the problem of vanishing gradients \citep{Veit16Residual}.
Yet, analyzing deep networks rather than shallow ones has remained a major challenge.

Recently, \citet{jacot2018neural} have shown that, under certain conditions (one of them is strong over-parameterization), training a deep neural network with gradient descent can be characterized by kernel regression with the neural tangent kernel (NTK). 
\tomt{
Essentially, this approach can be understood as a linearization (first-order Taylor series expansion) of the network's output with respect to its parameters around the initialization. 
It should be noted that the NTK is not the nominal regime of the non-linear learning capabilities of deep neural networks (e.g., like classical kernels \citep{scholkopf2002learning}, its feature mapping does not adapt to the data and 
it has not been shown to reach the performance of powerful deep networks \citep{chizat2018note}). 
Yet, 
the NTK can be used to 
identify or provide tractable analyses for
phenomena that are also observed in other deep learning settings}, such as achieving zero training loss \citep{jacot2018neural,chizat2018note,lee2019wide,arora2019exact} and faster learning of lower frequencies \citep{ronen2019convergence}.

As the NTK formulas depend on the network architecture, 
most of the NTK works consider the classical multilayer perceptron (MLP), a plain feed-forward network with fully connected layers 
\citep{jacot2018neural,chizat2018note,lee2019wide,arora2019fine,ronen2019convergence,bietti2019inductive,williams2019gradient, 
geifman2020similarity,chen2020deep,bietti2020deep}. 
\tomtr{Yet, some} 
recent papers compute 
\tomtr{NTK expressions} 
for other architectures \citep{arora2019exact,yang2019scaling,yang2020tensor,huang2020deep,alemohammad2020recurrent,hron2020infinite}.

{\bf Contribution.} In this paper, we develop the NTK for a ResNet model. 
After obtaining the formulas for the infinite width limit at initialization, we prove the stability of empirical NTK during training with gradient descent (and other common NTK assumptions), which implies that the trained ResNet model is indeed characterized by its NTK.
Note that proving stability during training is the key result that allows NTK-based analysis of neural networks. Yet, it is missing in a recent paper \citep{huang2020deep} that also considered a resembling ResNet model (more details are in Section~\ref{sec:ntk_resnet}).

By comparing the ResNet NTK and the MLP NTK 
\tomtb{for ReLU activations,} 
we find that ResNet promotes smoother interpolations than MLP \tomt{(without using any explicit regularization)}, which adds to other advantages of ResNet described in previous works. 
Our smoothness findings are based on different evaluation methodologies, such as visualizing the kernel of each model (which is data-independent), comparing \tomtb{uniform} upper bounds on the norm of the models' Jacobians after training, and comparing kernel regression results (specifically, interpolations) for the different NTKs. 
\tomtb{In the latter methodology, we use 
approximate 
$\mathcal{L}^2$-norm of the second derivative of a function as a quantitative measure for its smoothness.} 
Finally, we show that 
\tomtb{for ReLU-based networks} 
the smoothness advantage of ResNet is also 
observed 
outside the NTK regime.

Our analysis may 
\tomtr{be related to} 
the better generalization ability of ResNets over MLP, 
\tomt{as there is prior work that connects fitting the training data with a smoother function to better generalization error \tomtr{(under some smoothness assumption on the target function)} \citep{lu2019deeponet,giryes2020function,xie2020weighted}.}
We also show that the smoothness distinction between the two models can be increased by moderately attenuating the residual blocks when summing them with the skip connections.
Indeed, this practice has been shown to improve the training and generalization robustness of ResNets in a recent empirical classification study \citep{zhang2019towards}
\tomt{and its follow-up work \citep{yang2019dynamical}}.

\section{Background and Related Work}
\label{sec:prelim}

This section presents the NTK
of a plain MLP 
with some of its results. 
Consider an MLP model with $L$ hidden layers, input $\x \in \mathbb{R}^{d}$, parameter vector $\btheta := \mathrm{vec}(\{\W^{(\ell)}\})$, and output $\f(\x; \btheta) \in \mathbb{R}^k$, given by
\begin{align}
\label{Eq_mlp}
\g^{(\ell)} &= \frac{\sigma_w}{\sqrt{n_{\ell-1}}} \W^{(\ell)} \x^{(\ell-1)},\hspace{5mm} \ell=1,\ldots,L \\ \nonumber
\x^{(\ell)} &= \phi(\g^{(\ell)}),\hspace{5mm} \ell=1,\ldots,L \\ \nonumber
\x^{(0)} &=\x, \hspace{5mm} \f(\x,\btheta) = \g^{(L+1)} = \frac{\sigma_w}{\sqrt{n_{L}}} \W^{(L+1)} \x^{(L)},
\end{align}
where $\phi(\cdot)$ is an element-wise activation function, 
\tomtb{$\sigma_w$ is a positive hyperparameter that scales the standard deviation of $\{ \W^{(\ell)} \}$,}  
$\W^{(\ell)} \in \mathbb{R}^{n_{\ell} \times n_{\ell-1}}$, $n_0=d$, $n_{L+1}=k$, and 
all the weights are initialized by the standard normal distribution $W^{(\ell)}_{ij} \sim \mathcal{N}(0,1)$. 
It is assumed that the input is bounded $\|\x\|_2 \leq B$.

At initialization, when $n_1,\ldots,n_L \xrightarrow{} \infty$ 
each pre-activation $g^{(\ell)}_i(\x)$, and thus also $f_i(\x) = g^{(L+1)}_i(\x)$,
is a stochastic Gaussian Process (GP) with zero mean \citep{neal2012bayesian,lee2017deep}.
Denote the GP kernel (covariance) of this process by $K^{(L+1)}(\x,\tilde{\x}) := \Exppp{\btheta}{g^{(L+1)}_i(\x) g^{(L+1)}_i(\tilde{\x})}$ (note its independence of the entry index $i$). We have that $$f_i(\x) f_i(\tilde{\x}) \xrightarrow{n_{1:L} \xrightarrow{} \infty} K^{(L+1)}(\x,\tilde{\x}),$$ 
\tomt{where the limit should be interpreted in the almost surely sense.} 
Since $\Exppp{\btheta}{g^{(L+1)}_i(\x) g^{(L+1)}_j(\tilde{\x})}=0$ for $i\neq j$, for multidimensional output, we simply have  $$\f(\x) \f^\top(\tilde{\x}) \xrightarrow{n_{1:L} \xrightarrow{} \infty} K^{(L+1)}(\x,\tilde{\x}) \otimes \I_k,$$ where $\otimes$ is the Kronecker product. 
The GP kernel of the MLP above can be computed using the following recursive expression \citep{jacot2018neural,lee2017deep,yang2019wide}
\begin{align}
\label{Eq_mlp_gp}
K^{(L+1)}(\x,\tilde{\x}) &= \sigma_w^2 T \left (  \left  [ \begin{matrix} K^{(L)}(\x,\x) & K^{(L)}(\x,\tilde{\x}) \\
  K^{(L)}(\x,\tilde{\x}) & K^{(L)}(\tilde{\x},\tilde{\x}) \\
\end{matrix} \right  ]  \right ), \\ \nonumber
K^{(1)}(\x,\tilde{\x}) &= \frac{\sigma_w^2}{d}\x^\top\tilde{\x},
\end{align}
where $T(\bSigma):=\Exppp{ ( u,v )  \sim \mathcal{N}(\0,\bSigma)}{\phi(u) \phi(v)}$. 

Recently, a new type of kernel has received much attention, namely, the NTK \citep{jacot2018neural}, which is defined as $\Theta^{(L+1)}(\x,\tilde{\x}) := \Exp_{\btheta} \left \langle \frac{\partial  f_i(\x;\btheta)}{\partial \btheta} , \frac{\partial  f_i(\tilde{\x};\btheta)}{\partial \btheta} \right \rangle$ where $n_{1:L} \xrightarrow{} \infty$. 
Similarly to the GP kernel, it can be shown that at initialization $\left \langle \frac{\partial  f_i(\x;\btheta)}{\partial \btheta} , \frac{\partial  f_i(\tilde{\x};\btheta)}{\partial \btheta} \right \rangle \xrightarrow{n_{1:L} \xrightarrow{} \infty} \Theta^{(L+1)}(\x,\tilde{\x})$ \citep{jacot2018neural,arora2019exact,yang2019scaling,yang2020tensor}. (The extension to multidimensional output is done again by Kronecker product with $\I_k$).
Note that the significant impact of NTK is mainly due to the fact that it can be used to characterize DNN training with gradient flow or gradient descent (with small enough learning rate) \citep{jacot2018neural,lee2019wide,arora2019exact}.

Specifically, given the training data $\mathcal{D} = (\mathcal{X} , \mathcal{Y})$ 
\tomtb{(where $(\mathcal{X} , \mathcal{Y})=((\x_1,\y_1),\ldots,(\x_{|\mathcal{X}|},\y_{|\mathcal{X}|}))$ are the training samples $\{\x_i\} \in \mathbb{R}^d$ and their associated labels $\{\y_i\} \in \mathbb{R}^k$)} 
and a loss function $\ell(\cdot,\cdot): \mathbb{R}^k \times \mathbb{R}^k \to \mathbb{R}$, consider learning $\btheta$ by minimizing the empirical loss $\mathcal{L}=\sum_{(\x_i,\y_i)\in\mathcal{D}}\ell(\f(\x_i;\btheta),\y_i)$ using gradient descent with learning rate $\eta$. 
This can be written in continuous time (for simplicity) as 
$\dot{\btheta}_t = -\eta \frac{\partial \f(\mathcal{X};\btheta_t)}{\partial \btheta}^\top \nabla_{\f(\mathcal{X};\btheta_t)}\mathcal{L}$, 
where $\f(\mathcal{X};\btheta_t)=\mathrm{vec}(\{ \f(\x_i;\btheta_t) \}_{\x_i\in\mathcal{X}}) \in \mathbb{R}^{k|\mathcal{X}| \times 1}$.
In the function space, we have
\begin{align}
\label{Eq_gd_dynamic}
\dot{\f}(\mathcal{X};\btheta_t) &= \frac{\partial \f(\mathcal{X};\btheta_t)}{\partial \btheta} \dot{\btheta}_t  \\ \nonumber
&= - \eta \frac{\partial \f(\mathcal{X};\btheta_t)}{\partial \btheta} \frac{\partial \f(\mathcal{X};\btheta_t)}{\partial \btheta}^\top \nabla_{\f(\mathcal{X};\btheta_t)}\mathcal{L}.
\end{align}
Under appropriate conditions, it can be shown that 
$\frac{\partial \f(\mathcal{X};\btheta_t)}{\partial \btheta} \frac{\partial \f(\mathcal{X};\btheta_t)}{\partial \btheta}^\top \xrightarrow{n_{1:L} \xrightarrow{} \infty} \bTheta \otimes \I_k$, where $\bTheta \in \mathbb{R}^{|\mathcal{X}| \times |\mathcal{X}|}$ with $\Theta_{ij}=\Theta^{(L+1)}(\x_i,\x_j)$.
In other words, the dynamics of the output function in \eqref{Eq_gd_dynamic} turns into a simple ODE with respect to $\f(\mathcal{X};\btheta_t)$  
based on the constant NTK. \tomtr{For squared $\ell_2$ loss this is even a linear ODE with a close-from solution.} 
In this case, fitting scalar labels $\{y_i\}$ 
is reduced to 
$\ell_2$ kernel regression, which has the following closed-form solution  
\begin{align}
\label{Eq_kernel_sol}
f(\x)=\k(\x)^\top \bTheta^{-1}\y,
\end{align}
where $k_i(\x)=\Theta^{(L+1)}(\x,\x_i)$ and $\y=[y_1,\ldots,y_{|\mathcal{D}|}]^\top$.
\tomt{
A more general study on linear behaviour of non-linear models, which takes into account under-parameterized models and the effect of the scaling used in initialization, appears in \citep{chizat2018note}.}

Finally, the NTK of the MLP model in \eqref{Eq_mlp} can be computed using the following recursive expression \citep{jacot2018neural}
\begin{align}
\label{Eq_mlp_ntk}
\Theta^{(L+1)}(\x,\tilde{\x}) 
&= K^{(L+1)}(\x,\tilde{\x}) 
+ \Theta^{(L)}(\x,\tilde{\x}) \cdot \sigma_w^2 \dot{T} \left (  \left  [ \begin{matrix} K^{(L)}(\x,\x) & K^{(L)}(\x,\tilde{\x}) \\
  K^{(L)}(\x,\tilde{\x}) & K^{(L)}(\tilde{\x},\tilde{\x}) \\
\end{matrix} \right  ]  \right ), \\ \nonumber
\Theta^{(1)}(\x,\tilde{\x}) &= K^{(1)}(\x,\tilde{\x}),
\end{align}
where $\dot{T}(\bSigma):=\Exppp{ ( u,v )  \sim \mathcal{N}(\0,\bSigma)}{\phi'(u) \phi'(v)}$. 
Note that $T(\bSigma)$ and $\dot{T}(\bSigma)$ have closed-form expressions for the ReLU and erf activation functions. \tomt{For completeness, their expressions for ReLU, which are due to \citep{cho2009kernel}, are provided in Appendix~D. 
}

\section{NTK for ResNet}
\label{sec:ntk_resnet}

We turn now to develop the ResNet NTK. Consider a ResNet model with $L$ non-linear hidden layers, input $\x \in \mathbb{R}^{d}$, parameter vector $\btheta := \mathrm{vec}(\w^{(L+1)},\{\W^{(\ell)}\},\{\V^{(\ell)}\},\U)$, and output $f(\x; \btheta) \in \mathbb{R}$ given by
\begin{align}
\label{Eq_resnet}
\g^{(\ell)} &= \frac{\sigma_w}{\sqrt{n}} \W^{(\ell)} \x^{(\ell-1)},\hspace{5mm} \ell=1,\ldots,L \\ \nonumber
\x^{(\ell)} &= \x^{(\ell-1)} + \alpha \frac{\sigma_v}{\sqrt{n}} \V^{(\ell)} \phi(\g^{(\ell)}),\hspace{5mm} \ell=1,\ldots,L \\ \nonumber
\x^{(0)} &=\frac{1}{\sqrt{d}}\U\x, \hspace{5mm} f(\x,\btheta) = g^{(L+1)} = \frac{\sigma_w}{\sqrt{n}} \w^{(L+1)\top} \x^{(L)},
\end{align}
where $\phi(\cdot)$ is an element-wise activation function, 
\tomtb{$\sigma_w$ and $\sigma_v$ are positive hyperparameters that scale the standard deviation of $\{ \W^{(\ell)} \}$ and $\{ \V^{(\ell)} \}$, respectively, 
and $\alpha$ is a positive hyperparameter that weighs the residual block.} 
$\W^{(\ell)},\V^{(\ell)} \in \mathbb{R}^{n \times n}$, $\w^{(L+1)} \in \mathbb{R}^n$, $\U \in \mathbb{R}^{n \times d}$, and 
all the weights are initialized by the standard normal distribution $w^{(L+1)}_{i},W^{(\ell)}_{ij},V^{(\ell)}_{ij},U_{ij} \sim \mathcal{N}(0,1)$. 
It is assumed that the input is bounded $\|\x\|_2 \leq B$.

A few remarks are in place. First, we assume scalar output for simplification, and the extension to multidimensional output is straightforward as shown in Section~\ref{sec:prelim}. Second, note that we lift the input dimension from $\mathbb{R}^d$ to $\mathbb{R}^n$ using $\x^{(0)}=\frac{1}{\sqrt{d}}\U\x$. This lifting is unavoidable as the NTK analysis requires that the width of all intermediate layers approaches infinity. 
\tomtr{
Third, the weights $\{\V^{(\ell)}\}$ are necessary for the proof technique and for obtaining closed-form expressions for the ResNet NTK. Specifically, $\V^{(\ell)}$ breaks the correlation between $\x^{(\ell-1)}$ and $\phi(\g^{(\ell)})$ which leads to (relatively nice) formulas for the GP kernel and NTK with closed-form expressions in the case of ReLU activations. Also, as explained in Appendix~A, 
the multiplication of $\phi(\g^{(\ell)})$ by $\V^{(\ell)}$ allows us to use general convergence results from \citep{yang2019wide,yang2019scaling}. 
In contrast, ResNet models that do not multiply the nonlinear activations by $\{\V^{(\ell)}\}$ get complicated recursive equations for their kernel with {\em no closed-form analytical expressions even for ReLU activations} \citep{du2019gradient}, or do not provide kernel expressions at all \citep{zhang2019training}. 
Thus, they have no possibility to observe the shape of this kernel and obtain its kernel regression results (contrary to our ResNet NTK).} 

Lastly, a similar ResNet model is considered by \citep{huang2020deep}. 
\tomt{Yet, they assume that the weights of the first and last layers are fixed,} 
and they do not include a proof that when training this ResNet model with gradient descent/flow the limiting NTK stays the same as the one in the initialization (i.e., $\frac{\partial \f(\mathcal{X};\btheta_t)}{\partial \btheta} \frac{\partial \f(\mathcal{X};\btheta_t)}{\partial \btheta}^\top \xrightarrow{n \xrightarrow{} \infty} \bTheta$). 
Note that this missing result (proven here 
\tomt{in Theorem~\ref{thm:ntk_resnet_train}, using Lemma~\ref{thm:local_lip}})
is perhaps the most important property of NTK-based analysis of neural networks. 
\tomt{
Another difference between the works is that 
we use the derived NTK to compare the ResNet with MLP in terms of smoothness for a similar {\em fixed} depth (i.e., fixed $L$), while
\citet{huang2020deep} examine the NTKs expressions for $L \xrightarrow{} \infty$. In fact, note that the NTK regime for MLP requires  $L$ to be finite 
\citep{hanin2019finite,littwin2020residual}. 
}

Denote the {\em empirical} (random, finite-width) GP kernel and NTK at initialization by $\hat{K}_0^{(L+1)}(\x,\tilde{\x}) := f(\x;\btheta_0) f(\tilde{\x};\btheta_0)$ and $\hat{\Theta}^{(L+1)}_0(\x,\tilde{\x}) := \left \langle \frac{\partial  f(\x;\btheta_0)}{\partial \btheta} , \frac{\partial  f(\tilde{\x};\btheta_0)}{\partial \btheta} \right \rangle$, respectively.
Our first results state the GP kernel and NTK at initialization when $n \xrightarrow{} \infty$.
\tomtb{While these results are asymptotic, in Section~\ref{sec:ntk_resnet_vs_mlp} we present numerical experiments where the behavior of practical finite width networks correlates with the asymptotic NTK analysis. 
Yet, non-asymptotic results may be obtained using concentration bounds, as done for other kernel approximation techniques \citep{sriperumbudur2015optimal,koppel2019parsimonious}.}

\begin{theorem}[GP kernel at initialization]
\label{thm:gp_resnet}
Consider the ResNet model in \eqref{Eq_resnet}. We have $\hat{K}_0^{(L+1)}(\x,\tilde{\x})$ $\xrightarrow{n \xrightarrow{} \infty} K^{(L+1)}(\x,\tilde{\x}) := \Exppp{\btheta}{f(\x;\btheta) f(\tilde{\x};\btheta)}$, where 
$K^{(L+1)}(\x,\tilde{\x})$ can be computed recursively as following:
\begin{align}
\label{Eq_resnet_gp}
K^{(L+1)}(\x,\tilde{\x}) 
&= K^{(L)}(\x,\tilde{\x}) 
+ \alpha^2 \sigma_v^2 \sigma_w^2 T \left (  \left  [ \begin{matrix} K^{(L)}(\x,\x) & K^{(L)}(\x,\tilde{\x}) \\
  K^{(L)}(\x,\tilde{\x}) & K^{(L)}(\tilde{\x},\tilde{\x}) \\
\end{matrix} \right  ]  \right ), \\ \nonumber
K^{(1)}(\x,\tilde{\x}) &= \frac{\sigma_w^2}{d}\x^\top\tilde{\x}.
\end{align}
\end{theorem}

\begin{theorem}[NTK at initialization]
\label{thm:ntk_resnet}
Consider the ResNet model in \eqref{Eq_resnet} 
and let the element-wise non-linearities be bounded uniformly by $\mathrm{e}^{(cx^2-\epsilon)}$ for some $c,\epsilon>0$.
We have that $\hat{\Theta}_0^{(L+1)}(\x,\tilde{\x}) \xrightarrow{n \xrightarrow{} \infty} \Theta^{(L+1)}(\x,\tilde{\x}) := \Exp_{\btheta} \left \langle \frac{\partial  f(\x;\btheta)}{\partial \btheta} , \frac{\partial  f(\tilde{\x};\btheta)}{\partial \btheta} \right \rangle$, where $\Theta^{(L+1)}(\x,\tilde{\x})$ is given by
\begin{align}
\label{Eq_resnet_ntk}
\Theta^{(L+1)}(\x,\tilde{\x}) &= K^{(L+1)}(\x,\tilde{\x}) + \Pi^{(0)}(\x,\tilde{\x}) \cdot K^{(1)}(\x,\tilde{\x}) \nonumber \\ 
&\hspace{3mm} 
+ \alpha^2 \sum \limits_{\ell=1}^{L} \Pi^{(\ell)}(\x,\tilde{\x}) 
\cdot \left ( \Sigma^{(\ell+1)}(\x,\tilde{\x}) + K^{(\ell)}(\x,\tilde{\x}) \cdot \dot{\Sigma}^{(\ell+1)}(\x,\tilde{\x})  \right )
\end{align}
such that 
\begin{align}
\label{Eq_resnet_ntk_def}
\Sigma^{(\ell+1)}(\x,\tilde{\x}) := \sigma_v^2 \sigma_w^2 T \left (  \left  [ \begin{matrix} K^{(\ell)}(\x,\x) & K^{(\ell)}(\x,\tilde{\x}) \\
  K^{(\ell)}(\x,\tilde{\x}) & K^{(\ell)}(\tilde{\x},\tilde{\x}) \\
\end{matrix} \right  ]  \right ), \nonumber \\
\dot{\Sigma}^{(\ell+1)}(\x,\tilde{\x}) := \sigma_v^2 \sigma_w^2 \dot{T} \left (  \left  [ \begin{matrix} K^{(\ell)}(\x,\x) & K^{(\ell)}(\x,\tilde{\x}) \\
  K^{(\ell)}(\x,\tilde{\x}) & K^{(\ell)}(\tilde{\x},\tilde{\x}) \\
\end{matrix} \right  ]  \right ), 
\end{align}
$\{ K^{(\ell)}(\x,\tilde{\x}) \}$ are given in \eqref{Eq_resnet_gp}, 
and $\{ \Pi^{(\ell)}(\x,\tilde{\x}) \}$ can be computed using the following recursive expression
\begin{align}
\label{Eq_resnet_ntk_def_pi}
\Pi^{(\ell)}(\x,\tilde{\x}) &= \Pi^{(\ell+1)}(\x,\tilde{\x}) \left (  1 + \alpha^2 \dot{\Sigma}^{(\ell+2)}(\x,\tilde{\x})  \right ), \\ \nonumber
\Pi^{(L)}(\x,\tilde{\x}) &= 1.
\end{align}
\end{theorem}

The proofs of the theorems can be found in Appendix~A. 
\tomt{
Theorems~\ref{thm:gp_resnet} and \ref{thm:ntk_resnet} provide kernels that are associated with the considered ResNet model. Both of these kernels can be used in applications of kernel methods \citep{scholkopf2002learning}.
Yet, in what follows we show that the special property of the NTK holds also for our ResNet. 
Namely, that
under appropriate conditions, 
the limiting NTK stays 
constant 
even during gradient descent training of the ResNet. Therefore, as discussed below \eqref{Eq_gd_dynamic}, the network function during and after training can be characterized by kernel regression with the NTK.
}

Let $\btheta_t$ denote the parameters at time step $t$. Given training data $\mathcal{D} = (\mathcal{X} , \mathcal{Y})$ 
\tomtb{(where $(\mathcal{X} , \mathcal{Y})=((\x_1,y_1),\ldots,(\x_{|\mathcal{X}|},y_{|\mathcal{X}|}))$ are the training samples $\{\x_i\} \in \mathbb{R}^d$ and their associated labels $\{y_i\} \in \mathbb{R}$)}, 
we make the following shorthand notations
\begin{align}
\label{Eq_def_shorthands}
\f(\btheta_t) &= \mathrm{vec}(\{ f(\x_i;\btheta_t) \}_{\x_i\in\mathcal{X}}) \in \mathbb{R}^{|\mathcal{X}|}, \\ \nonumber
\e(\btheta_t) &= \f(\btheta_t) - \mathcal{Y} \in \mathbb{R}^{|\mathcal{X}|}, \\ \nonumber
\J(\btheta_t) &= \frac{\partial \f(\btheta_t)}{\partial \btheta} \in \mathbb{R}^{|\mathcal{X}| \times |\btheta|}.
\end{align}
The empirical $|\mathcal{X}| \times |\mathcal{X}|$ NTK Gram matrix is defined as
\begin{align}
\label{Eq_def_grams}
\hat{\bTheta}_t := \hat{\bTheta}_t(\mathcal{X},\mathcal{X}) = \J(\btheta_t) \J(\btheta_t)^\top.
\end{align}
From Theorem~\ref{thm:ntk_resnet} we have that $\hat{\bTheta}_0 \xrightarrow{n \xrightarrow{} \infty} \bTheta \in \mathbb{R}^{|\mathcal{X}| \times |\mathcal{X}|}$ with $\Theta_{ij}=\Theta^{(L+1)}(\x_i,\x_j)$.

In Theorem~\ref{thm:ntk_resnet_train} below, we show that when training the ResNet using the loss function $\mathcal{L}(\btheta)=\frac{1}{2}\sum_{(\x_i,y_i)\in\mathcal{D}} (f(\x_i;\btheta) - y_i )^2 = \frac{1}{2} \| \e(\btheta) \|_2^2$ and gradient descent with small enough learning rate $\eta$, we get $\supr{t} \| \hat{\bTheta}_t - \hat{\bTheta}_0 \|_F = \mathcal{O}(\frac{1}{\sqrt{n}})$, which implies
$\hat{\bTheta}_t \xrightarrow{n \xrightarrow{} \infty} \bTheta$. 
To obtain the result $\supr{t} \| \hat{\bTheta}_t - \hat{\bTheta}_0 \|_F = \mathcal{O}(\frac{1}{\sqrt{n}})$ we extend the strategy of \citep{lee2019wide} from MLP to the ResNet.
The extension is based on the following two lemmas. 

\begin{lemma}
\label{thm:gaussian_lambda}
Let $\W$ be an $m \times n$ random matrix whose entries are independent standard normal variables. Then for every $t \geq 0$, with probability at least $1-2\mathrm{exp}(-t^2/2)$ we have
\begin{align}
\label{Eq_gaussian_lambda}
\sqrt{m} - \sqrt{n} - t \leq \lambda_{min}(\W) \leq \lambda_{max}(\W) \leq \sqrt{m} + \sqrt{n} + t,
\end{align}
where $\lambda_{min}(\W)$ and $\lambda_{max}(\W)$ denote the smallest and largest singular values of $\W$, respectively.
\end{lemma}

\begin{lemma}
\label{thm:local_lip}
Consider the ResNet model in \eqref{Eq_resnet} initialized with $\btheta_0$, and assume that the activation function $\phi$ satisfies $|\phi(z)| \leq C_\phi |z|$, $|\phi'(z)| \leq C_\phi$ and $|\phi(z)-\phi(\tilde{z})|, |\phi'(z)-\phi'(\tilde{z})| \leq C_\phi |z-\tilde{z}|$, for some $C_\phi>0$.
Then, there exists a $K>0$ (that does not depend on $n$) such that for every $C>0$ and $n \gg C^2$, with high probability over the random initialization, the following holds for all $\btheta,\tilde{\btheta} \in B(\btheta_0,C):=\{\btheta : \|\btheta-\btheta_0\|_2 \leq C\}$
\begin{align}
\label{Eq_local_lip}
&\| \J(\btheta) \|_F \leq K,  \\ \nonumber
&\| \J(\btheta) - \J(\tilde{\btheta}) \|_F \leq K \| \btheta - \tilde{\btheta} \|_2.
\end{align}
\end{lemma}

Lemma~\ref{thm:gaussian_lambda} is adopted from \citep{vershynin2010introduction} (Corollary 5.35 there).
It is used to prove Lemma~\ref{thm:local_lip} and will be used again later in this paper.
Lemma~\ref{thm:local_lip} extends the ``MLP version" that appears in \citep{lee2019wide} to the considered ResNet model. Yet, due to the structure of the ResNet that is much more complex than for MLP, the proof of Lemma~\ref{thm:local_lip} is more complex and is deferred to Appendix~B. 
%
Using Lemma~\ref{thm:local_lip}, we have the following theorem.

\begin{theorem}[Stability of the NTK during training]
\label{thm:ntk_resnet_train}
Consider the ResNet model in \eqref{Eq_resnet} with activation function that satisfies the conditions from Lemma~\ref{thm:local_lip}.  
Assume that $\lambda_{min}(\bTheta)>0$, the training set $\mathcal{D} = (\mathcal{X} , \mathcal{Y})$ is contained in some compact set and $\x \neq \tilde{\x}$ for all $\x,\tilde{\x} \in \mathcal{X}$. 
Then, for $\delta_0>0$ there exist $R_0>0$, $N$ and $K>1$, such that for every $n>N$ when applying gradient descent on $\mathcal{L}(\btheta)= \frac{1}{2} \| \e(\btheta) \|_2^2$ with learning rate $\eta_0 < 2(\lambda_{min}(\bTheta)+\lambda_{max}(\bTheta))^{-1}$ the following holds with probability at least $1-\delta_0$ over the random initialization
\begin{align}
\label{Eq_thm_ntk_resnet_train}
&\| \e(\btheta_t) \|_2 \leq \left (1-\frac{\eta_0}{3} \lambda_{min}(\bTheta) \right )^t R_0,  \\ \nonumber
&\sum \limits_{j=1}^t \| \btheta_j - \btheta_{j-1} \|_2 
\leq \frac{3K R_0}{\lambda_{min}(\bTheta)},  \\ \nonumber
&\sup_{t} \| \hat{\bTheta}_t - \hat{\bTheta}_0 \|_F = \frac{6K^3 R_0}{\lambda_{min}(\bTheta)}n^{-0.5}.
\end{align}
\end{theorem}

\begin{proof}
\tomtr{
The proof is based on induction and applying Lemma~\ref{thm:local_lip} with $C=\frac{3K R_0}{\lambda_{min}(\bTheta)}$. Essentially, it is an
extension of Theorems G.1 and G.4 in \citep{lee2019wide} from MLP to the considered ResNet model. Notice that while the proof of Lemma~\ref{thm:local_lip} (local boundness and Lipschitzness of the gradient) is very different for different network architectures (e.g., ResNet), the other steps that are required to prove these theorems do not depend on the network model.}
\end{proof}

Note that the first line in \eqref{Eq_thm_ntk_resnet_train} implies convergence to zero training loss, the second line implies stability of the weights during training (the bound on their amount of change does not depend on the network width $n$), and the third line shows the stability of the NTK Gram matrix, which implies $\hat{\bTheta}_t \xrightarrow{n \xrightarrow{} \infty} \bTheta$, as discussed above.

\section{Comparing the Smoothness of ResNet and MLP NTKs}
\label{sec:ntk_resnet_vs_mlp}

\tomt{
In this section, we compare the smoothness of the results of ResNet and MLP in the NTK regime (and beyond) using different evaluation methodologies. 
The smoothness property of a learned function (especially, of an interpolation) is of 
\tomtr{interest also because}
prior works have connected it with better generalization \citep{lu2019deeponet,giryes2020function,xie2020weighted}.} 
We start with comparing \tomtb{uniform (i.e., for any input)}  
upper bounds on 
the norm of the models' Jacobians {\em after training}, which is possible due to the NTK regime. 
This analysis formally shows that decreasing $\alpha$ in the ResNet model limits the non-smoothness of the learned function 
and reduces its associated bound below the bound obtained for MLP. 
Therefore, we examine different values of $\alpha$ also in other evaluation methodologies, such as kernel visualization and measuring the smoothness of NTK regression outputs by 
an approximated 
$\mathcal{L}^2$-norm of the outputs' second derivatives.

\tomtb{
Throughout this section, we focus on the distinction between MLPs and ResNets for ReLU activations, which are extremely popular in practice, and for which $T(\K)$ and $\dot{T}(\K)$ in the GP kernel and the NTK have closed-form expressions (see Appendix~D).
The smoothness distinction may not exist in other cases, such as when the activations are the identity function (then, the networks merely learn linear maps), or when the activations are the smooth erf function (see Appendix~E).}

\subsection{Comparing Bounds on Models' Jacobians}
\label{sec:ntk_resnet_vs_mlp__jecobians}

In the NTK regime, i.e., 
when the conditions of Theorem~\ref{thm:ntk_resnet_train} hold, we get from the second line in \eqref{Eq_thm_ntk_resnet_train} that for any $t$ we have $\btheta_t \in B(\btheta_0, \frac{3K R_0}{\lambda_{min}(\bTheta)})$, which by Lemma~\ref{thm:gaussian_lambda} implies that for $\sqrt{n} \gg  \frac{3K R_0}{\lambda_{min}(\bTheta)}$ the parameters in the NTK regime are tightly connected to their Gaussian initialization.
Formally, recall that $\btheta_0 = \mathrm{vec}(\w_0^{(L+1)},\{\W_0^{(\ell)}\},\{\V_0^{(\ell)}\},\U_0)$, where all the elements in $\btheta_0$ are i.i.d.~standard normal.
Let $\btheta \in B(\btheta_0,C)$. Therefore, the spectral norm of $\W^{(\ell)}$ obeys
\begin{align}
\label{Eq_W_bound2}   
\| \W^{(\ell)} \| &\leq \| \W_0^{(\ell)} \| + \| \W^{(\ell)} - \W_0^{(\ell)} \| \\ \nonumber
&\leq 2\sqrt{n} + t + C \leq 3\sqrt{n},
\end{align}
where the first inequality uses the triangular inequality, the second inequality uses Lemma~\ref{thm:gaussian_lambda} and $\| \W^{(\ell)} - \W_0^{(\ell)} \| \leq \| \W^{(\ell)} - \W_0^{(\ell)} \|_F \leq \|\btheta-\btheta_0\|_2 \leq C$, and the last inequality holds with high probability
 for $\sqrt{n} \gg C$.
Using the same arguments we have $\| \V^{(\ell)} \| \leq 3\sqrt{n}$, $\| \U \| \leq \sqrt{d} + 2\sqrt{n}$ and $\| \w^{(L+1)} \|_2 \leq 2\sqrt{n}$.
This mean that, in the NTK regime (of both ResNet and MLP) the spectral norm of the weights can be easily bounded. 
This is in contrast with the general case where  
there is no convenient way to control the weights of DNNs {\em after} training.

Using these properties of the NTK regime  
for {\em finite}, yet large $n$ 
(namely, 
$\sqrt{n} \gg  \frac{3K R_0}{\lambda_{min}(\bTheta)}$), 
we show   
the benefit of using small values for the hyperparameter $\alpha$ in ResNets. 
 The advantage of this setting has been empirically demonstrated for classification by \citet{zhang2019towards} (outside the NTK regime). 
\tomtb{
An indicator of the smoothness of a (trained) network $f(\x)$, which is also amenable to analysis, 
can be the maximal norm of the network's ``input-output Jacobian'' $\supr{\x} \left \| \frac{\partial}{\partial \x}f(\x) \right \|_2$ (similarly to the way Lipschitz continuity of the gradient is used in the optimization literature).  Smaller upper bound on this quantity can be interpreted as higher smoothness.}

Considering the ResNet model in \eqref{Eq_resnet}, we have 
\begin{align}
\label{Eq_resent_jacobian}    
\frac{\partial}{\partial \x}f_{ResNet}(\x) 
&= \frac{\partial f }{\partial \x^{(L)}} \frac{\partial \x^{(L)} }{\partial \x^{(L-1)}} \ldots   \frac{\partial \x^{(1)} }{\partial \x^{(0)}} \frac{\partial \x^{(0)} }{\partial \x}  \\ \nonumber
&= \frac{\sigma_w}{\sqrt{n}}\w^{(L+1)\top} 
\Bigg (  \prod \limits_{\ell=1}^{L}  \Big (   \I_n 
+ \alpha \frac{\sigma_v}{\sqrt{n}} \V^{(\ell)} \mathrm{diag} \left \{ \phi'(\g^{(\ell)}) \right \} \frac{\sigma_w}{\sqrt{n}} \W^{(\ell)}  \Big )  \Bigg )  \frac{1}{\sqrt{d}}\U.
\end{align}
Let us bound 
\tomtb{$\supr{\x} \left \|\frac{\partial}{\partial \x}f_{ResNet}(\x) \right \|_2$}
\begin{align}
\label{Eq_resent_jacobian2_}    
\supr{\x} \left \|\frac{\partial}{\partial \x}f_{ResNet}(\x) \right \|_2
&\leq \frac{\sigma_w}{\sqrt{n}} \| \w^{(L+1)} \|_2 \frac{1}{\sqrt{d}} \| \U \| \cdot 
\prod \limits_{\ell=1}^{L} \left ( 1 + \alpha C_\phi \frac{\sigma_v}{\sqrt{n}} \| \V^{(\ell)} \|  \frac{\sigma_w}{\sqrt{n}} \| \W^{(\ell)} \|  \right ) \\ \nonumber
& \leq \frac{\sigma_w}{\sqrt{n}} 2\sqrt{n} \frac{1}{\sqrt{d}} (\sqrt{d}+2\sqrt{n})\cdot 
\prod \limits_{\ell=1}^{L} \left ( 1 + \alpha C_\phi \frac{\sigma_v}{\sqrt{n}} 3\sqrt{n}  \frac{\sigma_w}{\sqrt{n}} 3\sqrt{n}  \right ) \\ \nonumber
& \leq 2\sigma_w (1+ 2\sqrt{\frac{n}{d}}) \left ( 1 + 9 \alpha C_\phi \sigma_v \sigma_w  \right )^L := B_{\mathrm{ResNet}}.
\end{align}
It 
can be seen that a smaller value of $\alpha$ decreases the bound, 
which hints that it encourages 
 $f_{\mathrm{ResNet}}(\x)$ to be smoother.
Note also that $\left \| \frac{\partial}{\partial \x}f_{ResNet}(\x) \right \| = \mathcal{O}(\sqrt{n})$ (which we got due to the fact that, as done in all NTK models, the weights $\U$ that are applied on the input are normalized by the input dimension $\frac{1}{\sqrt{d}}$ rather than by $\frac{1}{\sqrt{n}}$). 
This factor is not surprising, since we proved that under rather mild conditions on $\mathcal{X}$, the ResNet can fit any training data in the NTK regime, and thus the slope of $f_{\mathrm{ResNet}}(\x)$ is not bounded by a constant number.
Therefore, to obtain a more formal result on the advantage of small $\alpha$, let us relate the above result to the one that is obtained for MLP, which also has the same $\sqrt{\frac{n}{d}}$ factor because of the normalization of the first layer. 

%
Considering the MLP model in \eqref{Eq_mlp} with $n_1=\ldots=n_L=n$ and scalar output, we get the following input-output Jacobian 
\begin{align}
\label{Eq_mlp_jacobian}    
\frac{\partial}{\partial \x} f_{MLP}(\x)
&= \frac{\partial f }{\partial \x^{(L)}} \frac{\partial \x^{(L)} }{\partial \x^{(L-1)}} \ldots   \frac{\partial \x^{(1)} }{\partial \x^{(0)}} \frac{\partial \x^{(0)} }{\partial \x}  \\ \nonumber
&= \frac{\sigma_w}{\sqrt{n}}\w^{(L+1)\top} \left (  \prod \limits_{\ell=2}^{L} \mathrm{diag} \left \{ \phi'(\g^{(\ell)}) \right \} \frac{\sigma_w}{\sqrt{n}} \W^{(\ell)}  \right ) 
\mathrm{diag} \left \{ \phi'(\g^{(1)}) \right \} \frac{\sigma_w}{\sqrt{d}} \W^{(1)}.
\end{align}
Let us bound \tomtb{$\supr{\x} \left \|\frac{\partial}{\partial \x}f_{MLP}(\x) \right \|_2$} 
(recall that $\W^{(1)} \in \mathbb{R}^{n \times d}$ in the MLP)
\begin{align}
\label{Eq_mlp_jacobian2_}    
\supr{\x} \left \|\frac{\partial}{\partial \x}f_{MLP}(\x) \right \|_2
&\leq \frac{\sigma_w}{\sqrt{n}} \| \w^{(L+1)} \|_2 C_\phi \frac{\sigma_w}{\sqrt{d}} \|\W^{(1)}\|  \prod \limits_{\ell=2}^{L}  C_\phi  \frac{\sigma_w}{\sqrt{n}} \| \W^{(\ell)} \|  \\ \nonumber
& \leq \frac{\sigma_w}{\sqrt{n}} 2\sqrt{n} C_\phi \frac{\sigma_w}{\sqrt{d}} (\sqrt{d}+2\sqrt{n})  \prod \limits_{\ell=2}^{L}  C_\phi  \frac{\sigma_w}{\sqrt{n}} 3\sqrt{n}  \\ \nonumber
& \leq 2 C_\phi \sigma_w^2  (1+ 2\sqrt{\frac{n}{d}})  \left ( 3 C_\phi \sigma_w \right )^{L-1} := B_{\mathrm{MLP}}.
\end{align}

\begin{figure}[t]
  \centering
  \subfigure[{\scriptsize Empirical and asymptotic NTK}]{%
    \label{fig:NTK_regression_asymp_empir_kernel}
    \includegraphics[width=140pt]{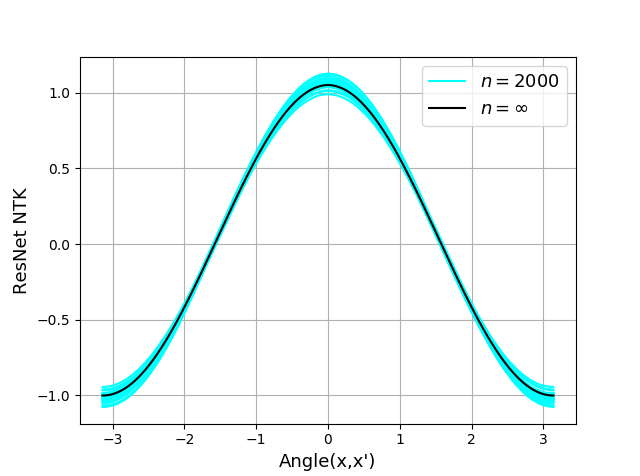}
  }
  \subfigure[{\scriptsize Interpolation with 6 samples}]{%
    \label{fig:NTK_regression_asymp_empir_regression_N6}
    \includegraphics[width=140pt]{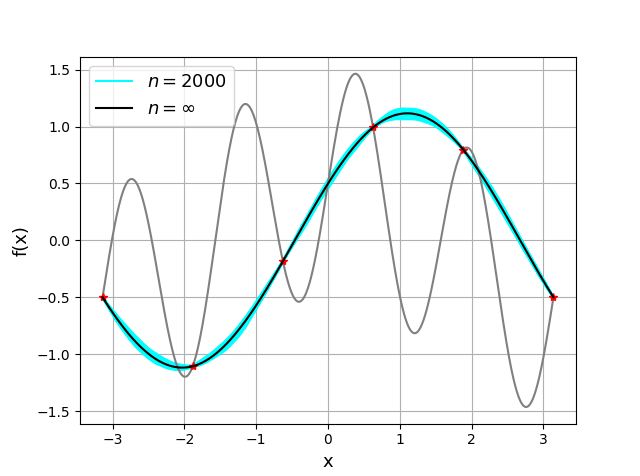}
  }
  \subfigure[{\scriptsize Interpolation with 10 samples}]{%
    \label{fig:NTK_regression_asymp_empir_regression_N10}
    \includegraphics[width=140pt]{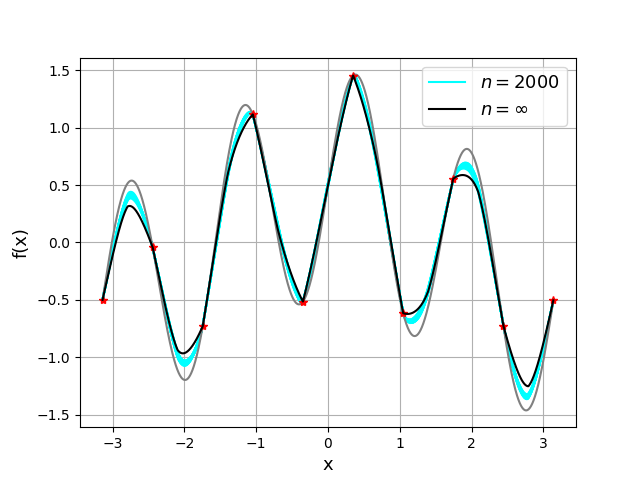}
  }

    \caption{Empirical (finite width of $n=2000$ and 30 different Gaussian initializations) and asymptotic NTK for ResNet with $L=5$ nonlinear layers, ReLU nonlinearities, $\alpha=0.1$, $\sigma_v=\sigma_w=1$, for inputs on the sphere (circle) in $\mathbb{R}^2$.
    (\ref{fig:NTK_regression_asymp_empir_kernel}): The kernel shape.  
    (\ref{fig:NTK_regression_asymp_empir_regression_N6})-(\ref{fig:NTK_regression_asymp_empir_regression_N10}): Interpolation using the closed-form NTK solution and using gradient descent training of the finite-width ResNet (5K iterations with lr 0.05 in (\ref{fig:NTK_regression_asymp_empir_regression_N6}), and 10K iterations with lr 0.5 in (\ref{fig:NTK_regression_asymp_empir_regression_N10})).  
    }
\label{fig:NTK_asymp_empir}     
\end{figure}

Comparing \eqref{Eq_resent_jacobian2_} and \eqref{Eq_mlp_jacobian2_}, we can compute the value of $\alpha$ for which $B_{\mathrm{ResNet}} \leq B_{\mathrm{MLP}}$. 
\tomtb{
We assume that the value of the constants $\sigma_v,\sigma_w$ and $C_\phi$ is 1, as common in practice.
Specifically, $C_\phi$ bounds the expansiveness of the activation function $\phi(\cdot)$ and its derivative (see Lemma~\ref{thm:local_lip}), and equals 1 for the widely used ReLU activation. 
The hyperparameters $\sigma_v$ and $\sigma_w$ scale the weight matrices, and setting $\sigma_v=\sigma_w=1$ coincides for square matrices with the popular Xavier’s Gaussian initialization (where the standard deviation of the entries is $1/\sqrt{n}$, as appears in our models).} 
Thus, we get
\begin{align}
\label{Eq_jacobian_comp}  
\frac{B_{\mathrm{ResNet}}}{B_{\mathrm{MLP}}} &= \frac{ ( 1 + 9 \alpha C_\phi \sigma_v \sigma_w )^L}{3^{L-1}(  C_\phi \sigma_w)^{L}} = \frac{(1+9\alpha)^L}{3^{L-1}} \\ \nonumber
&\leq 1 \iff \alpha \leq \frac{ 3^{1-1/L}-1}{9}
\end{align}
We can see that a moderate value of $\alpha$, such as 0.1, implies $B_{\mathrm{ResNet}} \leq B_{\mathrm{MLP}}$ for any $L \geq 3$. Interestingly, this is the fine-tuned value that has been used in the empirical classification paper by \citet{zhang2019towards}. 
Note 
that for $\alpha \xrightarrow{} 0$ the residual blocks cannot be trained. Thus, using too small $\alpha$ is not recommended in practice. Even in the NTK regime we have observed no 
advantage in extremely low $\alpha$.
\tomtb{
Finally, \eqref{Eq_jacobian_comp} hints that if $\alpha$ is small (e.g., smaller than 0.1) then increasing the number of nonlinear layers $L$ will increase the smoothness advantage of the ResNet NTK. This behavior is also demonstrated empirically in the sequel. 
}

\begin{figure}[t]
  \centering
  \subfigure[{\scriptsize NTKs (normalized to unit peak) $L=5$}]{%
    \label{fig:NTK_mlp_resnet_L5_kernels}
    \includegraphics[width=140pt]{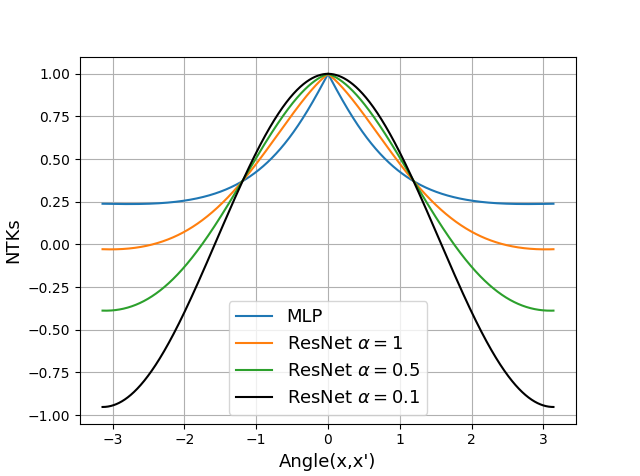}
  }
  \subfigure[{\scriptsize Interpolation with 6 samples $L=5$}]{%
    \label{fig:NTK_mlp_resnet_L5_regression_N6}
    \includegraphics[width=140pt]{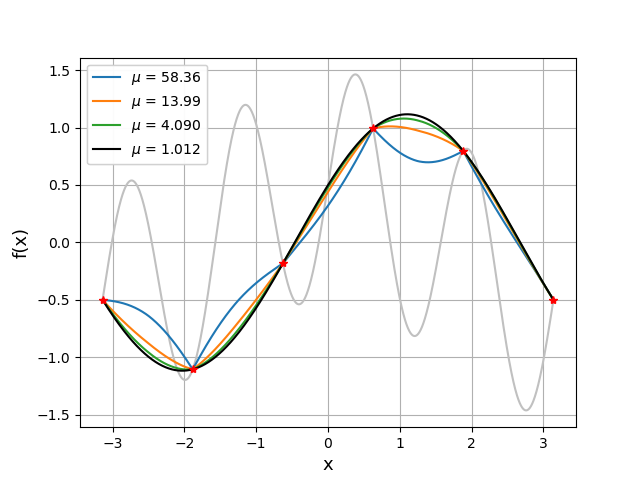}
  }
  \subfigure[{\scriptsize Interpolation with 10 samples $L=5$}]{%
    \label{fig:NTK_mlp_resnet_L5_regression_N10}
    \includegraphics[width=140pt]{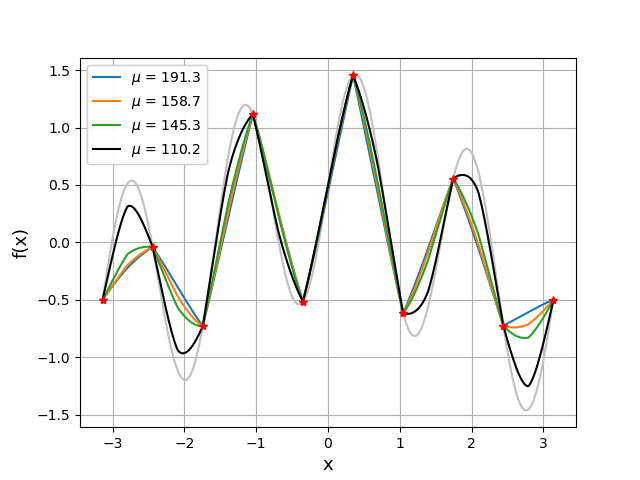}
  }



  \subfigure[{\scriptsize NTKs (normalized to unit peak) $L=15$}]{%
    \label{fig:NTK_mlp_resnet_L15_kernels}
    \includegraphics[width=140pt]{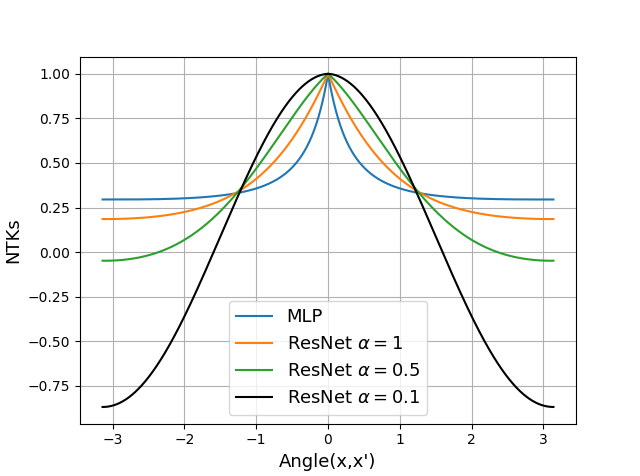}
  }
  \subfigure[{\scriptsize Interpolation with 6 samples $L=15$}]{%
    \label{fig:NTK_mlp_resnet_L15_regression_N6}
    \includegraphics[width=140pt]{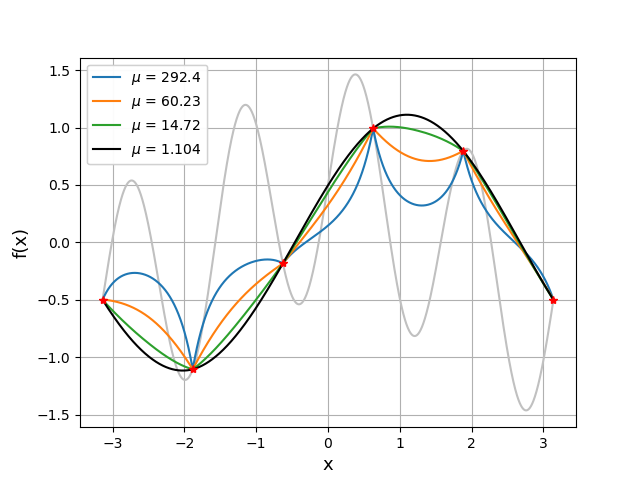}
  }
  \subfigure[{\scriptsize Interpolation with 10 samples $L=15$}]{%
    \label{fig:NTK_mlp_resnet_L15_regression_N10}
    \includegraphics[width=140pt]{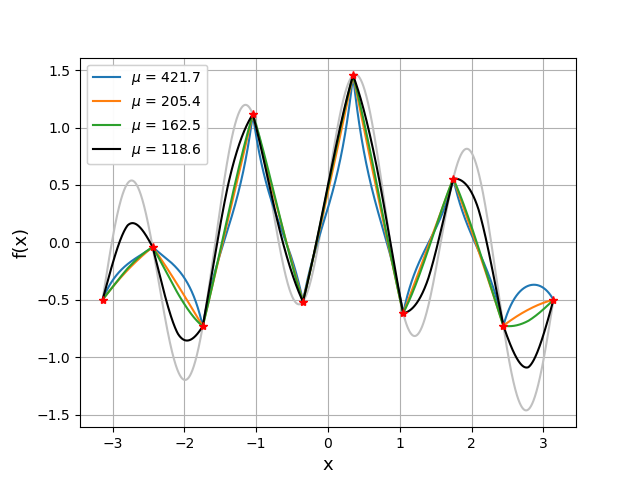}
  }

    \caption{NTKs for MLP and ResNet (for different values of $\alpha$) with $L=5$ (top) and $L=15$ (bottom) nonlinear layers, ReLU nonlinearities, $\sigma_v=\sigma_w=1$, for inputs on the sphere (circle) in $\mathbb{R}^2$.
    (\ref{fig:NTK_mlp_resnet_L5_kernels}),(\ref{fig:NTK_mlp_resnet_L15_kernels}): The kernels shape.  
    (\ref{fig:NTK_mlp_resnet_L5_regression_N6})-(\ref{fig:NTK_mlp_resnet_L5_regression_N10}), (\ref{fig:NTK_mlp_resnet_L15_regression_N6})-(\ref{fig:NTK_mlp_resnet_L15_regression_N10}): Interpolations by the closed-form solutions, measured by $\mu(\cdot)$ defined in \eqref{Eq_L2_der2_meas2}. 
    Note that the legend in (\ref{fig:NTK_mlp_resnet_L5_kernels}),(\ref{fig:NTK_mlp_resnet_L15_kernels}) applies to all the figures. 
    }
\label{fig:NTK_mlp_resnet_L15}     
\end{figure}

\subsection{Comparing the Kernels and their Interpolations}
\label{sec:ntk_resnet_vs_mlp__regression}

In this section, we compare the smoothness of the ResNet NTK and the MLP NTK by visualizing the kernel of each model (which is data-independent if the input norm is fixed), and by comparing the interpolations obtained by kernel regression with the different NTKs. 
\tomtb{
For a given interpolation $f$ with a scalar input $x \in [-\pi,\pi]$ and a scalar output, 
we scale it to unit $\mathcal{L}^2$-norm, $\bar{f}:=f/\|f\|_{\mathcal{L}^2}$, and
use 
an approximate 
$\mathcal{L}^2$-norm of the second derivative of $\bar{f}$ 
as a quantitative measure of smoothness  
(where a smaller value is interpreted as higher smoothness).
We empirically find this measure to be richer than first-derivative quantities, like $\mathrm{sup}_{x}|f'(x)|$ 
that is more tractable for analysis (as we have done above) but 
is affected by linear slopes of $f$ and cannot sum the effects of multiple nonsmooth points.}

\tomtb{
We numerically approximate $\|\bar{f}''\|_{\mathcal{L}^2}$ as follows. 
We densely sample $f$ at $N=4096$ ``test samples" with equal spacing $\Delta x = \frac{2\pi}{N}$, i.e., at $x_q=q\Delta x$ with $q=-N/2,...,N/2-1$. 
We compute its normalized version $\bar{f}=f/\widehat{\|f\|_{\mathcal{L}^2}}$, where  $\widehat{\|f\|_{\mathcal{L}^2}}=(\frac{1}{2\pi} \sum_q |f(x_q)|^2 \Delta x)^{0.5}$.
As the interpolations in our experiments are periodic ($f(-\pi)=f(\pi)$), we utilize the Fourier series representation of $\bar{f}$ to mitigate numerical computation issues of discrete derivatives. 
We approximate the $k$th Fourier coefficient $c_k=\frac{1}{2\pi} \int_{-\pi}^{\pi} \bar{f}(x) \mathrm{e}^{-jkx} dx$ by $\frac{1}{2\pi} \sum_{q=-N/2}^{N/2-1} \bar{f}(x_q) \mathrm{e}^{-jk q 2 \pi/N} \Delta x = \frac{1}{N} \mathrm{FFT}[ \{ \bar{f}(x_q) \} ](k)$.
Thus, denoting $\mathrm{FFT}[ \{ \bar{f}(x_q) \} ](k)$ by $F(k)$,  
we approximate 
$\|\bar{f}''\|_{\mathcal{L}^2}$ by 
\begin{align}
\label{Eq_L2_der2_meas2}   
\mu(f):= \left ( \frac{1}{N^2} \sum_{k=-N/2}^{N/2-1} |k|^4 | F(k) |^2 \right )^{\frac{1}{2}},
\end{align}
where we used 
Parseval's identity and the property that 
the $k$th Fourier coefficient of the $r$th derivative of $\bar{f}$, $\bar{f}^{(r)}$, is given by $(jk)^r c_k$ (where $c_k$ is the $k$th Fourier coefficient of $\bar{f}$). 
}

\tomt{Note that scaling a kernel by a scalar factor does not change its kernel regression result $f$ and thus also $\mu(f)$. Therefore, we found the measure $\mu(f)$ to be more informative than, e.g., comparing the FFTs of the ResNet and MLP NTKs (we observed that with large/small enough factor the magnitude of the FFT of each of them can be placed on-top/below the other).}

\begin{figure}[t]
  \centering
  \subfigure[{\scriptsize Interpolation with 6 samples (Adam)}]{%
    \label{fig:outside_NTK_regime_N6_adam}
    \includegraphics[width=140pt]{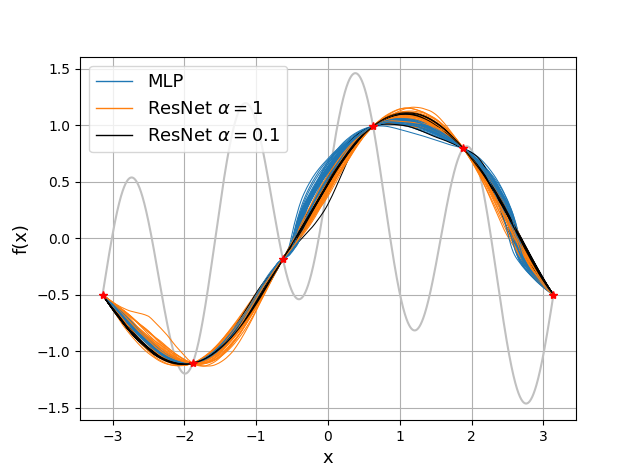}
  }
  \subfigure[{\scriptsize Interpolation with 6 samples (SGD)}]{%
    \label{fig:outside_NTK_regime_N6_sgd}
    \includegraphics[width=140pt]{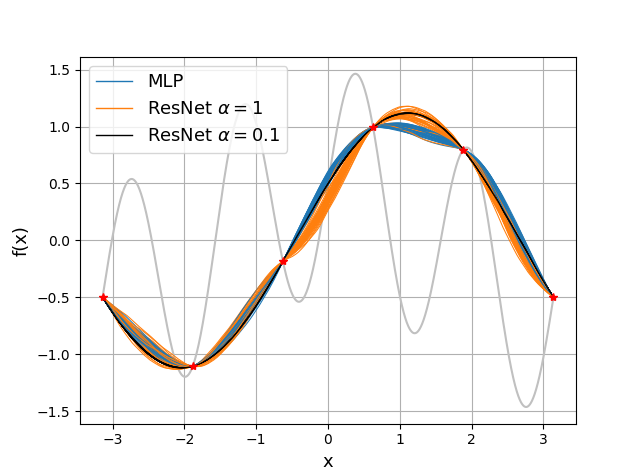}
  }
  \subfigure[{\scriptsize Interpolation with 10 samples (SGD)}]{%
    \label{fig:outside_NTK_regime_N10_sgd}
    \includegraphics[width=140pt]{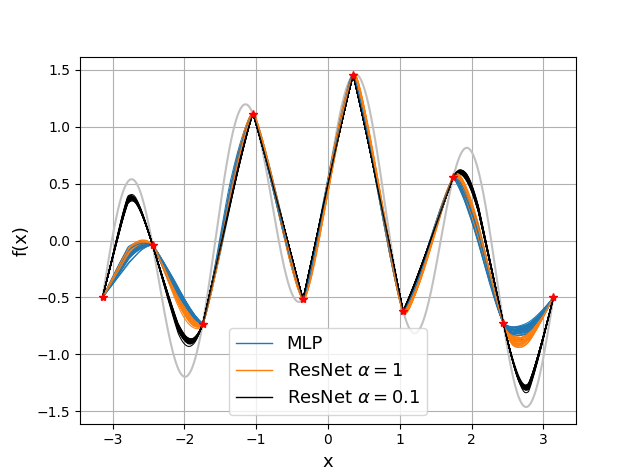}
  }

    \caption{Empirical 
    interpolations of MLP and ResNet (for different values of $\alpha$) with $L=5$ nonlinear layers, ReLU nonlinearities, $\sigma_v=\sigma_w=1$, for inputs on the sphere (circle) in $\mathbb{R}^2$. We use practical models with width of $n=500$, 30 different Xavier's Gaussian initializations (instead of normalizations by $1/\sqrt{n}$) and 1K iterations of SGD/Adam optimizers.
    }
\label{fig:outside_NTK_regime}     
\end{figure}

Before comparing with the MLP, we visualize the theoretical NTK results for ResNet (similar visualization of the NTK theory has been shown only for MLPs, e.g., in \citep{jacot2018neural,lee2019wide}).
Figure~\ref{fig:NTK_regression_asymp_empir_kernel} shows the concentration of the empirical NTK (for 30 different Gaussian initializations) around the asymptotic expression given in  \eqref{Eq_resnet_ntk}, for the ResNet model with $L=5$, ReLU nonlinearities, $\alpha=0.1$, $\sigma_v=1$, $\sigma_w=1$, and width of $n=2000$, for inputs on the sphere (circle) in $\mathbb{R}^2$.
Figures~\ref{fig:NTK_regression_asymp_empir_regression_N6} and \ref{fig:NTK_regression_asymp_empir_regression_N10} show 
that the results of kernel regression \eqref{Eq_kernel_sol} with the asymptotic NTK are very similar to the interpolations learned by the ResNet (for 30 different Gaussian initializations, when we use gradient descent with step-size 0.5 and 0.05 for 6 and 10 training samples, respectively).

We turn to compare the NTK results for MLP and ResNet, which are given in \eqref{Eq_mlp_ntk} and \eqref{Eq_resnet_ntk}, respectively. We use $\sigma_v=\sigma_w=1$, ReLU nonlinearities, and inputs from the circle. We modify the number of nonlinear layers, $L$, as well as the hyperparameter $\alpha$ in the ResNet. In the interpolation results (
where we use 
\eqref{Eq_kernel_sol}) we also modify the amount of given samples and measure the smoothness of each 
\tomtb{resulted $f$ with $\mu(f)$ defined in \eqref{Eq_L2_der2_meas2}.}  
The results are presented in 
Figure~\ref{fig:NTK_mlp_resnet_L15}. 

It can be seen that decreasing $\alpha$ yields a smoother ResNet NTK with smoother interpolation results. For $\alpha=1$ the ResNet NTK is more similar to the MLP NTK, but even then it appears smoother and less ``edgy" than the MLP. 
\tomtb{
Moreover, for the MLP NTK, increasing $L$ clearly reduces the smoothness (as observed both visually and by the increase in $\mu$).  For the ResNet NTK this effect is moderate for $\alpha=1$, and almost unseen for $\alpha=0.1$.}

\tomt{
More results and details are presented in Appendix~C, 
including showing that an extremely small value of $\alpha$ 
does not significantly affect the kernel shape and smoothness compared to the moderate $\alpha=0.1$. 
\tomtb{In Appendix~C.2  
we also present several visual results for a two-dimensional input and accuracy results for binary classification of high-dimensional input (MNIST data).}
}

\begin{figure}[t]
  \centering
  \subfigure[{\scriptsize Interpolation by MLP (Adam)}]{%
    \label{fig:outside_NTK_regime_appendix1_mlp_N20_adam}
    \includegraphics[width=140pt]{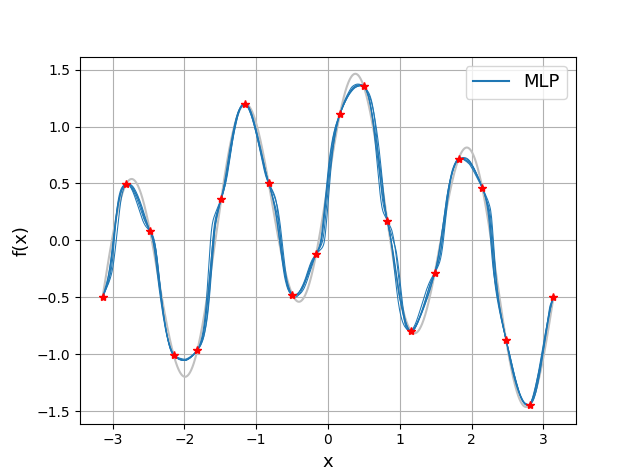}
  }
  \subfigure[{\scriptsize Interpolation by ResNet $\alpha=1$ (Adam)}]{%
    \label{fig:outside_NTK_regime_appendix1_resnet_1p0_N20_adam}
    \includegraphics[width=140pt]{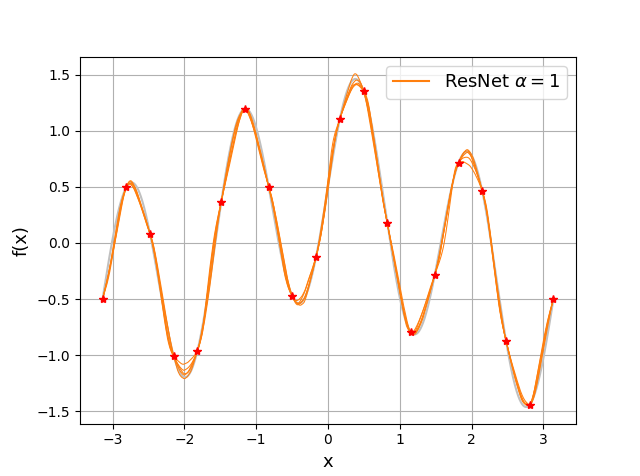}
  }
  \subfigure[{\scriptsize Interpolation by ResNet $\alpha=0.1$ (Adam)}]{%
    \label{fig:outside_NTK_regime_appendix1_resnet_0p1_N20_adam}
    \includegraphics[width=140pt]{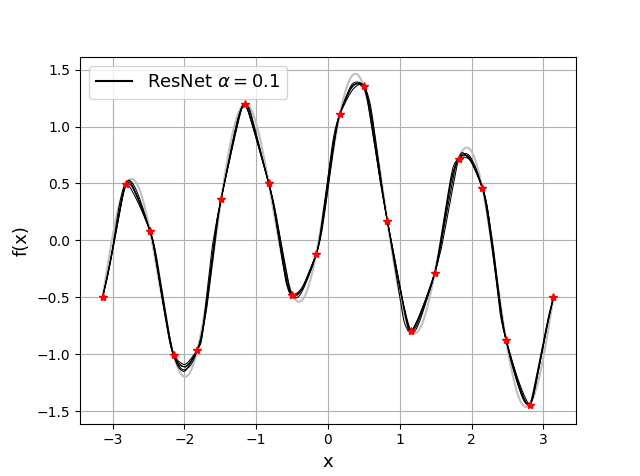}
  } 
   
    \caption{Empirical 
    interpolations of MLP and ResNet (for different values of $\alpha$) with $L=5$ nonlinear layers, ReLU nonlinearities, $\sigma_v=\sigma_w=1$, for inputs on the sphere (circle) in $\mathbb{R}^2$. We use practical models with width of $n=500$, 5 different Xavier's random Gaussian initializations (instead of normalizations by $1/\sqrt{n}$) and 1K iterations of Adam optimizer.
    }
\label{fig:outside_NTK_regime_appendix1}     

\vspace{5mm}

  \centering
  \subfigure[{\scriptsize Interpolation by MLP (Adam)}]{%
    \label{fig:outside_NTK_regime_appendix2_mlp_N20_adam}
    \includegraphics[width=140pt]{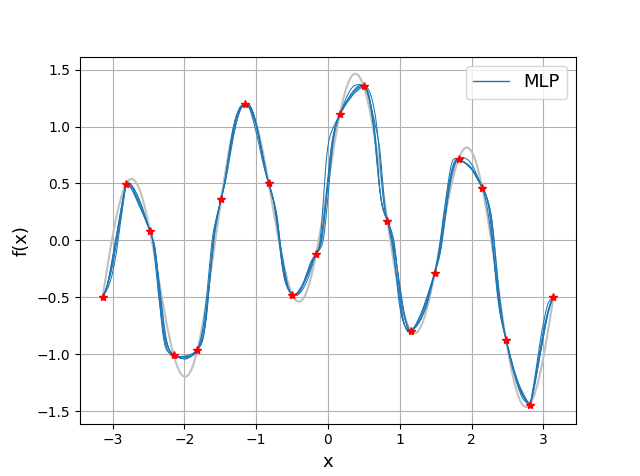}
  }
  \subfigure[{\scriptsize Interpolation by ResNet $\alpha=1$ (Adam)}]{%
    \label{fig:outside_NTK_regime_appendix2_resnet_1p0_N20_adam}
    \includegraphics[width=140pt]{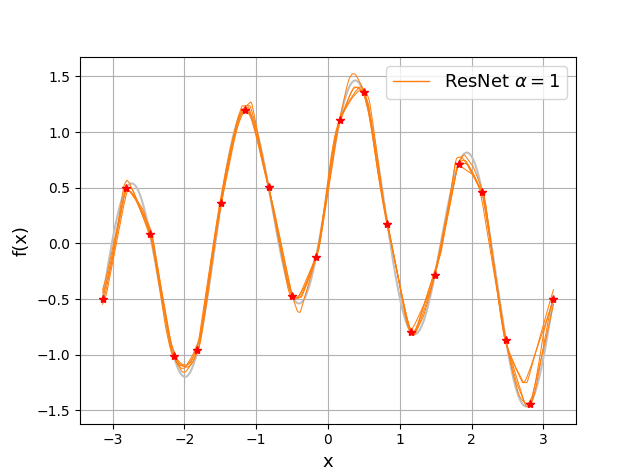}
  }
  \subfigure[{\scriptsize Interpolation by ResNet $\alpha=0.1$ (Adam)}]{%
    \label{fig:outside_NTK_regime_appendix2_resnet_0p1_N20_adam}
    \includegraphics[width=140pt]{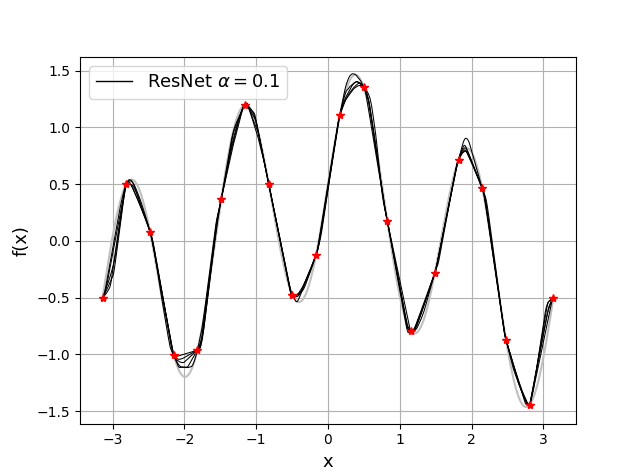}
  }     
   
    \caption{Empirical 
    interpolations of MLP and ResNet (for different values of $\alpha$) with $L=5$ nonlinear layers, ReLU nonlinearities, $\sigma_v=\sigma_w=1$, for scalar (1D) inputs. We use practical models with width of $n=500$ and biases, 5 different  PyTorch (default) random uniform initializations (instead of Gaussian initializations and normalizations by $1/\sqrt{n}$) and 1K iterations of Adam optimizer.
    }
\label{fig:outside_NTK_regime_appendix2}     
\end{figure}

\subsection{Results Outside the NTK Regime}
\label{sec:ntk_resnet_vs_mlp__outside}

Finally, we demonstrate that the observations that are made for the NTK regime carry also to other settings.
We consider MLP and ResNet with $L=5$ nonlinear layers and width of (only) $n=500$ neurons. We replace the NTK initializations (including the normalizations by $1/\sqrt{n}$) with Xavier's Gaussian initialization (this mainly affects the first and last layers), and instead of gradient descent we perform either SGD (with lr 0.01) or Adam (with the default parameters stated in \citep{kingma2014adam}) on ``mini-batches" of size 1. 
\tomt{
We emphasize that we do not use any explicit regularization (such as batch-normalization or weight decay).} 
The results for 30 different realizations of the initialization are presented in Figure~\ref{fig:outside_NTK_regime}.
\tomt{
}

Despite the discrepancy between these settings and the conditions that are required for the NTK regime to hold, we see similarity in the results. 
Even in these settings, it is clear that moderately 
decreasing $\alpha$ yields smoother interpolation results for the ResNet.
For $\alpha=0.1$ the results of ResNet are much smoother than those of MLP. 
For $\alpha=1$ the results of ResNet are more similar to the results of MLP, as observed the NTK regime. Yet, with Adam optimizer the results of ResNet are smoother also with $\alpha=1$.
As mentioned above, a detailed empirical study on the use of small values of $\alpha$ (outside the NTK Regime) for classification tasks has been done in \citep{zhang2019towards}.
Our work can be regarded as an NTK-based support for this approach.

\tomtr{
Next, we repeat the numerical experiments with 20 samples. 
In Figure~\ref{fig:outside_NTK_regime_appendix1} we present the interpolation results for different Xavier's Gaussian initializations, where the input to the networks is 2D points on the circle, as done in the previous experiments. The optimization method is 1K iterations of Adam with lr 1e-4 and ``mini-batches" of size 1, where we save the model with minimal training loss.} 

\tomtr{
In Figure~\ref{fig:outside_NTK_regime_appendix2} we take a step farther. We add bias to all the layers and feed the networks with plain scalar input. We replace the NTK initializations (including the normalizations by $1/\sqrt{n}$) with PyTorch default uniform initialization (which is similar to Kaiming's initialization). Again, we optimize by the Adam method, as mentioned above (recall that the theory requires gradient descent).}

\tomtr{
Despite the discrepancy between these settings and the conditions for the NTK regime, we see similarity in the results: 
The interpolations of the ResNets are smoother 
than those of the MLP.}

\section{Conclusion}

In this paper we developed the NTK for a ResNet model and proved its stability during training with gradient descent (under common NTK assumptions). 
As the smoothness of interpolators can indicate better generalization \citep{lu2019deeponet,giryes2020function,xie2020weighted}, we compared the smoothness properties of 
\tomtb{ReLU-based}  
ResNet and MLP in the regime where training them can be characterized by kernel regression with their associated NTKs. 
Our smoothness examination, which is based on different evaluation methodologies, shows that ResNet, especially with moderately attenuated residual blocks, yields smoother interpolations than MLP in the NTK regime. 
We also showed that this smoothness advantage of ResNet can be observed outside the NTK regime, i.e., 
when the settings differ from 
the NTK assumptions.

Our NTK analysis has captured the advantage of reducing the skip weighting factor $\alpha$. 
One may inquire whether it is possible 
to use the NTK regime for finding other new improvements to ResNet.

\acks{TT and RG acknowledge support from the European research council (ERC StG 757497 PI Giryes) and Nvidia for donating a GPU. 
JB acknowledges partial support from the Alfred P. Sloan Foundation, NSF RI-1816753, NSF CAREER CIF 1845360, and Samsung Electronics.}

\bibliography{msml_final}


\appendix

\numberwithin{equation}{section}

\newpage

\section{Proofs for Theorem~\ref{thm:gp_resnet} and Theorem~\ref{thm:ntk_resnet}}
\label{app:ntk_comp}

Note that the structure of the ResNet model in \eqref{Eq_resnet} shares similarities with the plain MLP model in \eqref{Eq_mlp}. For example, due to the central limit theorem when $n \xrightarrow{} \infty$ each pre-activation $g^{(\ell)}_i(\x)$ is a stochastic Gaussian Process (GP) with zero mean and deterministic GP kernel (covariance), just like in MLP.
Indeed, extension of the ``convergence at initialization" results of GP kernel and NTK for models beyond MLP has been shown in several works.

Specifically, the convergence of the GP kernel and NTK of the ResNet model that is considered in this paper follows from the general results of \citep{yang2019wide,yang2019scaling}. 
This can be done because our ResNet model follows the NETSOR approach \citep{yang2019wide,yang2019scaling}: 
It is built from {\em A-vars} (iid Gaussian weights distributed as $\mathcal{N}(0,\frac{\sigma_a^2}{n})$ for some $\sigma_a$), {\em g-vars} (Gussian vectors with iid entries given by multiplication of A-var with h-var or by sum of other g-vars) and {\em h-vars} (element-wise nonlinearity, bounded uniformly by $\mathrm{e}^{(cx^2-\epsilon)}$ for some $c,\epsilon>0$, applied on g-vars). 
For example, in our model 
$\x^{(0)}$ is g-var as the input can be considered as h-var, and then recursively 
$\x^{(\ell)}$ is g-var as it equals g-var + A-var $\times$ h-var. 
Therefore, it remains to compute the limiting kernels. 

\subsection{Computing the GP kernel for ResNet}
\label{app:ntk_comp_1}

To simplify the notation, we add the tilde symbol above each term that depends on the input $\tilde{\x}$ (rather than on $\x$), e.g., $\tilde{\x}^{(\ell)}$ denotes $\x^{(\ell)}(\tilde{\x})$. 
We will repeatedly use ``total expectation”, 
both for eliminating the cross-terms in $\Exp \left \langle \x^{(\ell)} , \tilde{\x}^{(\ell)} \right \rangle$, i.e., for $\Exp \left \langle \x^{(\ell-1)} , \V^{(\ell)} \phi(\tilde{\g}^{(\ell)} \right \rangle = 0$, as well as for exploiting $\Expp{\V^{(\ell)\top}\V^{(\ell)}}=\Expp{\W^{(\ell)\top}\W^{(\ell)}}=n\I_n$ and $\Expp{\w^{(L+1)}\w^{(L+1)\top}}=\I_n$.

First, note the identity
\begin{align}
\label{Eq_resent_gpk_comp0}    
\Expp{g_i^{(\ell)}  \tilde{g}_i^{(\ell)}} = \Expp{  \frac{\sigma_w}{\sqrt{n}} \x^{(\ell-1)\top} \w_i^{(\ell)}  \frac{\sigma_w}{\sqrt{n}} \w_i^{(\ell)\top} \tilde{\x}^{(\ell-1)} } 
= \frac{\sigma_w^2}{n} \Exp \left \langle \x^{(\ell-1)} , \tilde{\x}^{(\ell-1)} \right \rangle,
\end{align}
where $\w_i^{(\ell)\top}$ denotes the $i$th row of $\W^{(\ell)}$. 
Therefore, we have
\begin{align}
\label{Eq_resent_gpk_comp}    
K^{(L+1)}(\x,\tilde{\x}) = 
\Expp{g^{(L+1)}  \tilde{g}^{(L+1)}} 
= \frac{\sigma_w^2}{n} \Exp \left \langle \x^{(L)} , \tilde{\x}^{(L)} \right \rangle.
\end{align}
Using $\x^{(\ell)} = \x^{(\ell-1)} + \alpha \frac{\sigma_v}{\sqrt{n}} \V^{(\ell)} \phi(\g^{(\ell)})$ and $\Exp \left \langle \x^{(\ell-1)} , \V^{(\ell)} \phi(\tilde{\g}^{(\ell)} \right \rangle = 0$, we get
\begin{align}
\label{Eq_resent_gpk_comp2}    
K^{(L+1)}(\x,\tilde{\x}) &= \frac{\sigma_w^2}{n} \Exp \left \langle \x^{(L-1)} , \tilde{\x}^{(L-1)} \right \rangle + \alpha^2 \frac{\sigma_w^2}{n} \frac{\sigma_v^2}{n} \Exp \left \langle  \V^{(L)} \phi(\g^{(L)}) ,  \V^{(L)} \phi(\tilde{\g}^{(L)}) \right \rangle \\ \nonumber
&= \frac{\sigma_w^2}{n} \Exp \left \langle \x^{(L-1)} , \tilde{\x}^{(L-1)} \right \rangle + \alpha^2 \frac{\sigma_w^2 \sigma_v^2}{n} \Exp \left \langle \phi(\g^{(L)}) ,  \phi(\tilde{\g}^{(L)}) \right \rangle \\ \nonumber
&= \Expp{g_i^{(L)}  \tilde{g}_i^{(L)}} + \alpha^2 \sigma_w^2 \sigma_v^2 \Expp{\phi(g_i^{(L)}) \phi(\tilde{g}_i^{(L)}) } \\ \nonumber
&= K^{(L)}(\x,\tilde{\x}) + \alpha^2 \sigma_v^2 \sigma_w^2 T \left (  \left  [ \begin{matrix} K^{(L)}(\x,\x) & K^{(L)}(\x,\tilde{\x}) \\
  K^{(L)}(\x,\tilde{\x}) & K^{(L)}(\tilde{\x},\tilde{\x}) \\
\end{matrix} \right  ]  \right ).
\end{align}
We also have
\begin{align}
\label{Eq_resent_gpk_comp3}    
K^{(1)}(\x,\tilde{\x}) = \frac{\sigma_w^2}{n} \Exp \left \langle \x^{(0)} , \tilde{\x}^{(0)} \right \rangle = \frac{\sigma_w^2}{n} \frac{1}{d} \Expp{ \x^\top \U^\top \U \tilde{\x}} = \frac{\sigma_w^2}{d}\x^\top\tilde{\x}.
\end{align}

\subsection{Computing the NTK for ResNet}
\label{app:ntk_comp2}

To simplify the notation, we add the tilde symbol above each term that depends on the input $\tilde{\x}$ (rather than on $\x$), e.g., $\tilde{\x}^{(\ell)}$ denotes $\x^{(\ell)}(\tilde{\x})$. 
Recall the parameter vector $\btheta = \mathrm{vec}(\w^{(L+1)},\{\W^{(\ell)}\},\{\V^{(\ell)}\},\U)$. Therefore, we have 
\begin{align}
\label{Eq_resent_ntk_comp0}    
&\Theta^{(L+1)}(\x,\tilde{\x}) = \Exp \left \langle \frac{\partial  f(\x;\btheta)}{\partial \btheta} , \frac{\partial  f(\tilde{\x};\btheta)}{\partial \btheta} \right \rangle \\ \nonumber
&= \Exp \left \langle \frac{\partial  f}{\partial \w^{(L+1)}} , \frac{\partial  \tilde{f}}{\partial \w^{(L+1)}} \right \rangle + \Exp \left \langle \frac{\partial  f}{\partial \U} , \frac{\partial  \tilde{f}}{\partial \U} \right \rangle + \sum \limits_{\ell=1}^{L} \Exp \left \langle \frac{\partial  f}{\partial \W^{(\ell)}} , \frac{\partial  \tilde{f}}{\partial \W^{(\ell)}} \right \rangle + \Exp \left \langle \frac{\partial  f}{\partial \V^{(\ell)}} , \frac{\partial  \tilde{f}}{\partial \V^{(\ell)}} \right \rangle.
\end{align}

Clearly, $\frac{\partial  f}{\partial \w^{(L+1)}} = \frac{\partial  g^{(L+1)}}{\partial \w^{(L+1)}} = \frac{\sigma_w}{\sqrt{n}}\x^{(L)\top}$.
To express the other derivatives let us define
\begin{align}
\label{Eq_resent_ntk_delta}    
\bdelta^{(\ell)} := \nabla_{\x^{(\ell)}}f = \left ( \frac{\partial  f}{\partial \x^{(\ell)}} \right )^\top
= \left ( \frac{\sigma_w}{\sqrt{n}}\w^{(L+1)\top} \frac{\partial \x^{(L)} }{\partial \x^{(L-1)}} \ldots   \frac{\partial \x^{(\ell+1)} }{\partial \x^{(\ell)}}  \right )^\top,
\end{align}
and note that from $\x^{(\ell)} = \x^{(\ell-1)} + \alpha \frac{\sigma_v}{\sqrt{n}} \V^{(\ell)} \phi(\g^{(\ell)}) = \x^{(\ell-1)} + \alpha \frac{\sigma_v}{\sqrt{n}} \V^{(\ell)} \phi( \frac{\sigma_w}{\sqrt{n}} \W^{(\ell)} \x^{(\ell-1)} )$ we have 
\begin{align}
\label{Eq_resent_ntk_dx}    
\frac{\partial \x^{(\ell)} }{\partial \x^{(\ell-1)}} = \I_n + \alpha \frac{\sigma_v}{\sqrt{n}} \V^{(\ell)} \mathrm{diag} \left \{ \phi'(\g^{(\ell)}) \right \} \frac{\sigma_w}{\sqrt{n}} \W^{(\ell)}.
\end{align}
Other necessary derivatives are given by
\begin{align}
\label{Eq_resent_ntk_derivatives}    
\frac{\partial x_i^{(\ell)} }{\partial \W^{(\ell)}} &= \sum \limits_{j=1}^n \frac{\partial x_i^{(\ell)}}{\partial \phi(g_j^{(\ell)}) } \frac{\partial \phi(g_j^{(\ell)}) }{\partial \W^{(\ell)}}
= \alpha \frac{\sigma_v}{\sqrt{n}} \sum \limits_{j=1}^n V_{ij}^{(\ell)} \phi'(g_j^{(\ell)}) \frac{\sigma_w}{\sqrt{n}} \left  [ \begin{matrix} \0 \\ \x^{(\ell-1)\top} \,\, \mathrm{at \,\, row} \,\, j \\
  \0 \\
\end{matrix} \right  ] \\ \nonumber
&= \alpha \frac{\sigma_v}{\sqrt{n}} \frac{\sigma_w}{\sqrt{n}} \left  [ \begin{matrix} V_{i1}^{(\ell)} \phi'(g_1^{(\ell)}) \x^{(\ell-1)\top} \\ \vdots \\
  V_{in}^{(\ell)} \phi'(g_n^{(\ell)}) \x^{(\ell-1)\top} \\
\end{matrix} \right  ]
= \alpha \frac{\sigma_v}{\sqrt{n}} \frac{\sigma_w}{\sqrt{n}} \mathrm{diag} \left \{ \phi'(\g^{(\ell)}) \right \} \v_i^{(\ell)} \x^{(\ell-1)\top},  \\ \nonumber
\frac{\partial x_i^{(\ell)} }{\partial \V^{(\ell)}} &= \alpha \frac{\sigma_v}{\sqrt{n}} \left  [ \begin{matrix} \0 \\ \phi(\g^{(\ell)})^\top \,\, \mathrm{at \,\, row} \,\, i \\
  \0 \\
\end{matrix} \right  ],
\end{align}
where $\v_i^{(\ell)\top}$ denotes the $i$th row of $\V^{(\ell)}$. 
This yields 
\begin{align}
\label{Eq_resent_ntk_derivatives2}    
\frac{\partial f }{\partial \W^{(\ell)}} &= \sum \limits_{i=1}^n \frac{\partial f }{\partial x_i^{(\ell)}} \frac{\partial x_i^{(\ell)}}{\partial \W^{(\ell)}}
= \alpha \frac{\sigma_v}{\sqrt{n}} \frac{\sigma_w}{\sqrt{n}} \mathrm{diag} \left \{ \phi'(\g^{(\ell)}) \right \} \V^{(\ell)\top} \bdelta^{(\ell)} \x^{(\ell-1)\top}, \\ \nonumber
\frac{\partial f }{\partial \V^{(\ell)}} &= \sum \limits_{i=1}^n \frac{\partial f }{\partial x_i^{(\ell)}} \frac{\partial x_i^{(\ell)}}{\partial \V^{(\ell)}}
= \alpha \frac{\sigma_v}{\sqrt{n}} \bdelta^{(\ell)} \phi(\g^{(\ell)})^\top, \\ \nonumber
\frac{\partial f }{\partial \U} &= \sum \limits_{i=1}^n \frac{\partial f }{\partial x_i^{(0)}} \frac{\partial x_i^{(0)}}{\partial \U}
= \frac{1}{\sqrt{d}} \bdelta^{(0)} \x^\top.
\end{align}
Now we can compute the required expectations by repeatedly using ``total expectation", conditioning (mainly) on random variables after the $\ell$th layer, and 
noting that since we consider $n \xrightarrow{} \infty$ the covariance of $g_i^{(\ell)}, \tilde{g}_i^{(\ell)}$ (or $\frac{\sigma_w^2}{n} \x^{(\ell-1)\top} \tilde{\x}^{(\ell-1)}$) converges to the deterministic $K^{(\ell)}(\x,\tilde{\x})$, 
so 
values of $T(\cdot)$ and $\dot{T}(\cdot)$ are deterministic and can be taken out of the expectations.
\begin{align}
\label{Eq_resent_ntk_exp_comp1}    
\Exp \left \langle \frac{\partial  f}{\partial \w^{(L+1)}} , \frac{\partial  \tilde{f}}{\partial \w^{(L+1)}} \right \rangle &= \frac{\sigma_w^2}{n} \Exp \left \langle  \x^{(L)} , \tilde{\x}^{(L)}  \right \rangle = K^{(L+1)}(\x,\tilde{\x}).
\end{align}

\begin{align}
\label{Eq_resent_ntk_exp_comp1b}  
\Exp \left \langle \frac{\partial  f}{\partial \U} , \frac{\partial  \tilde{f}}{\partial \U} \right \rangle &= 
\Expp{ \bdelta^{(0)\top} \tilde{\bdelta}^{(0)} } \cdot \frac{1}{d} \x^\top \tilde{\x} = \frac{1}{\sigma_w^2} \Expp{ \bdelta^{(0)\top} \tilde{\bdelta}^{(0)} } \cdot K^{(1)}(\x,\tilde{\x})
\end{align}

\begin{align}
\label{Eq_resent_ntk_exp_comp2}   
\Exp \left \langle \frac{\partial  f}{\partial \V^{(\ell)}} , \frac{\partial  \tilde{f}}{\partial \V^{(\ell)}} \right \rangle &= \alpha^2 \Expp{ \bdelta^{(\ell)\top} \tilde{\bdelta}^{(\ell)} \cdot \frac{\sigma_v^2}{n} \phi(\g^{(\ell)})^\top \phi(\tilde{\g}^{(\ell)})  } \\ \nonumber
&= \alpha^2 \Expp{ \bdelta^{(\ell)\top} \tilde{\bdelta}^{(\ell)} \cdot \frac{\sigma_v^2}{n} \sum \limits_{i=1}^n \phi(g_i^{(\ell)}) \phi(\tilde{g}_i^{(\ell)})  } \\ \nonumber
&= \alpha^2 \Expp{ \bdelta^{(\ell)\top} \tilde{\bdelta}^{(\ell)} } \cdot \sigma_v^2 T \left (  \left  [ \begin{matrix} K^{(\ell)}(\x,\x) & K^{(\ell)}(\x,\tilde{\x}) \\
  K^{(\ell)}(\x,\tilde{\x}) & K^{(\ell)}(\tilde{\x},\tilde{\x}) \\
\end{matrix} \right  ]  \right ).
\end{align}

\begin{align}
\label{Eq_resent_ntk_exp_comp3}   
\Exp \left \langle \frac{\partial  f}{\partial \W^{(\ell)}} , \frac{\partial  \tilde{f}}{\partial \W^{(\ell)}} \right \rangle &= \alpha^2 \Expp{ \frac{\sigma_w^2}{n} \x^{(\ell-1)\top} \tilde{\x}^{(\ell-1)} \cdot \bdelta^{(\ell)\top} \V^{(\ell)} \frac{\sigma_v^2}{n} \mathrm{diag} \left \{ \phi'(\g^{(\ell)}) \right \} \mathrm{diag} \left \{ \phi'(\tilde{\g}^{(\ell)}) \right \} \V^{(\ell)\top} \tilde{\bdelta}^{(\ell)}  } \nonumber \\ 
&= \alpha^2 \Expp{ \frac{\sigma_w^2}{n} \x^{(\ell-1)\top} \tilde{\x}^{(\ell-1)} \cdot \bdelta^{(\ell)\top} \left ( \frac{\sigma_v^2}{n} \sum \limits_{i=1}^n  \phi'(g_i^{(\ell)})  \phi'(\tilde{g}_i^{(\ell)}) \v_i^{(\ell)} \v_i^{(\ell)\top} \right ) \tilde{\bdelta}^{(\ell)}  } \nonumber \\ 
&= \alpha^2 \Expp{ \frac{\sigma_w^2}{n} \x^{(\ell-1)\top} \tilde{\x}^{(\ell-1)} \cdot \bdelta^{(\ell)\top} \tilde{\bdelta}^{(\ell)} \cdot \frac{\sigma_v^2}{n} \sum \limits_{i=1}^n  \phi'(g_i^{(\ell)})  \phi'(\tilde{g}_i^{(\ell)}) }  \nonumber \\ 
&= \alpha^2 K^{(\ell)}(\x,\tilde{\x}) \cdot \Expp{ \bdelta^{(\ell)\top} \tilde{\bdelta}^{(\ell)} } \cdot \sigma_v^2 \dot{T} \left (  \left  [ \begin{matrix} K^{(\ell)}(\x,\x) & K^{(\ell)}(\x,\tilde{\x}) \\
  K^{(\ell)}(\x,\tilde{\x}) & K^{(\ell)}(\tilde{\x},\tilde{\x}) \\
\end{matrix} \right  ]  \right ).
\end{align}

Let us now derive a recursive expression for $\Pi^{(\ell)}(\x,\tilde{\x}) := \frac{1}{\sigma_w^2} \Expp{ \bdelta^{(\ell)\top} \tilde{\bdelta}^{(\ell)} }$, using the relation $\bdelta^{(\ell)} = \left ( \frac{\partial  \x^{(\ell+1)}}{\partial \x^{(\ell)}} \right )^\top \bdelta^{(\ell+1)} = \left ( \I_n + \alpha  \frac{\sigma_w}{\sqrt{n}} \W^{(\ell+1)\top} \mathrm{diag} \left \{ \phi'(\g^{(\ell+1)}) \right \} \frac{\sigma_v}{\sqrt{n}} \V^{(\ell+1)\top} \right ) \bdelta^{(\ell+1)}$

\begin{align}
\label{Eq_resent_ntk_exp_comp4}   
&\Pi^{(\ell)}(\x,\tilde{\x}) = \frac{1}{\sigma_w^2} \Expp{ \bdelta^{(\ell)\top} \tilde{\bdelta}^{(\ell)} }  \\ \nonumber
&= \frac{1}{\sigma_w^2} \Expp{ \bdelta^{(\ell+1)\top} \left ( \I_n + \alpha^2 \frac{\sigma_v^2}{n} \V^{(\ell+1)}  \mathrm{diag} \left \{ \phi'(\g^{(\ell+1)}) \right \}  \frac{\sigma_w^2}{n} \W^{(\ell+1)} \W^{(\ell+1)\top}  \mathrm{diag} \left \{ \phi'(\tilde{\g}^{(\ell+1)}) \right \}  \V^{(\ell+1)\top}  \right ) \tilde{\bdelta}^{(\ell+1)}  } \\ \nonumber
&= \frac{1}{\sigma_w^2} \Expp{ \bdelta^{(\ell+1)\top} \left ( \I_n + \alpha^2 \frac{\sigma_v^2 \sigma_w^2}{n} \V^{(\ell+1)}  \mathrm{diag} \left \{ \phi'(\g^{(\ell+1)}) \right \}  \mathrm{diag} \left \{ \phi'(\tilde{\g}^{(\ell+1)}) \right \}  \V^{(\ell+1)\top}  \right ) \tilde{\bdelta}^{(\ell+1)}  } \\ \nonumber
&= \frac{1}{\sigma_w^2} \Expp{ \bdelta^{(\ell+1)\top} \left ( \I_n + \alpha^2 \left ( \frac{\sigma_v^2 \sigma_w^2}{n} \sum \limits_{i=1}^n  \phi'(g_i^{(\ell+1)})  \phi'(\tilde{g}_i^{(\ell+1)}) \v_i^{(\ell+1)} \v_i^{(\ell+1)\top} \right )  \right ) \tilde{\bdelta}^{(\ell+1)}  } \\ \nonumber
&= \frac{1}{\sigma_w^2} \Expp{ \bdelta^{(\ell+1)\top} \tilde{\bdelta}^{(\ell+1)} \cdot \left ( 1 + \alpha^2 \frac{\sigma_v^2 \sigma_w^2}{n} \sum \limits_{i=1}^n  \phi'(g_i^{(\ell+1)})  \phi'(\tilde{g}_i^{(\ell+1)}) \right )  } \\ \nonumber
&= \Pi^{(\ell+1)}(\x,\tilde{\x}) \cdot \left ( 1 + \alpha^2 \sigma_v^2 \sigma_w^2 \dot{T} \left (  \left  [ \begin{matrix} K^{(\ell+1)}(\x,\x) & K^{(\ell+1)}(\x,\tilde{\x}) \\
  K^{(\ell+1)}(\x,\tilde{\x}) & K^{(\ell+1)}(\tilde{\x},\tilde{\x}) \\
\end{matrix} \right  ]  \right ) \right ).
\end{align}
Note that the reasoning for the third equality (where total expectation is used to handle $\W^{(\ell+1)}\W^{(\ell+1)\top}$) is delicate, since $\W^{(\ell+1)}$ appears also in $\g^{(\ell+1)}$. 
This obstacle, which occurs in all NTK works, is handled by assuming that $\W^{(\ell+1)\top}$ used in backprop is independent from $\W^{(\ell+1)}$ in $\g^{(\ell+1)}$ that is used in the forward pass. This assumption has been justified in the limit $n \xrightarrow{} \infty$, as long as the last layer weight ($\w^{(L+1)}$) is sampled independently from other parameters and has zero mean \citep{arora2019exact,yang2019scaling}.

Finally, we compute the base case $\ell=L$
\begin{align}
\label{Eq_resent_ntk_exp_comp5}   
&\Pi^{(L)}(\x,\tilde{\x}) = \frac{1}{\sigma_w^2} \Expp{ \bdelta^{(L)\top} \tilde{\bdelta}^{(L)} } = \frac{1}{\sigma_w^2} \Expp{ \frac{\sigma_w}{\sqrt{n}}\w^{(L+1)\top} \frac{\sigma_w}{\sqrt{n}}\w^{(L+1)} } = 1.
\end{align}
Substituting equations \ref{Eq_resent_ntk_exp_comp1}--\ref{Eq_resent_ntk_exp_comp3} in \eqref{Eq_resent_ntk_comp0} and using the definitions of $\Pi^{(\ell)}(\x,\tilde{\x})$, $\Sigma^{(\ell+1)}(\x,\tilde{\x})$ and $\dot{\Sigma}^{(\ell+1)}(\x,\tilde{\x})$, we get the expression for the ResNet NTK that appears in \eqref{Eq_resnet_ntk}.

\newpage

\section{Proof for Lemma~\ref{thm:local_lip}}
\label{app:local_lip}

Recall that $\btheta_0 := \mathrm{vec}(\w_0^{(L+1)},\{\W_0^{(\ell)}\},\{\V_0^{(\ell)}\},\U_0)$, where all the elements in $\btheta_0$ are i.i.d. standard normal.
Let $\btheta \in B(\theta_0,C)$. Therefore, with high probability
\begin{align}
\label{Eq_W_bound}   
\| \W^{(\ell)} \| \leq \| \W_0^{(\ell)} \| + \| \W^{(\ell)} - \W_0^{(\ell)} \| \leq 2\sqrt{n} + t + C \leq 3\sqrt{n},
\end{align}
where the first inequality uses the triangular inequality, the second inequality uses Lemma~\ref{thm:gaussian_lambda} and $\| \W^{(\ell)} - \W_0^{(\ell)} \| \leq \| \W^{(\ell)} - \W_0^{(\ell)} \|_F \leq C$ and the last inequality uses $n \gg C^2$ and holds with high probability.
Using the same arguments we have $\| \V^{(\ell)} \| \leq 3\sqrt{n}$, $\| \U \| \leq \sqrt{d} + 2\sqrt{n}$ and $\| \w^{(L+1)} \|_2 \leq 2\sqrt{n}$.

Observe that
\begin{align}
\label{Eq_local_lip_comp0}   
\| \J(\btheta) \|_F^2  &= \sum \limits_{\x \in \mathcal{X}} \left ( \left \| \frac{\partial  f(\x;\theta)}{\partial \w^{(L+1)}} \right \|_2^2 
+ \left \| \frac{\partial  f(\x;\theta)}{\partial \U} \right \|_F^2
+ \sum \limits_{\ell=1}^{L} \left \| \frac{\partial  f(\x;\theta)}{\partial \W^{(\ell)}} \right \|_F^2 + \left \| \frac{\partial  f(\x;\theta)}{\partial \V^{(\ell)}} \right \|_F^2  \right ). 
\end{align}
Let us bound the terms in the sum. We will use the equality $\|\a\b^\top \|_F=\|\a\|_2\|\b\|_2$ and the derivatives that are obtained in Appendix~\ref{app:ntk_comp2}.
\begin{align}
\label{Eq_local_lip_comp01}   
\left \| \frac{\partial  f(\x;\theta)}{\partial \w^{(L+1)}} \right \|_2 = \frac{\sigma_w}{\sqrt{n}} \left \| \x^{(L)} \right \|_2.
\end{align}

\begin{align}
\label{Eq_local_lip_comp01b}   
\left \| \frac{\partial  f(\x;\theta)}{\partial \U} \right \|_F = \frac{1}{\sqrt{d}} \left \| \bdelta^{(0)} \x^\top \right \|_F = \frac{1}{\sqrt{d}} \left \| \bdelta^{(0)} \right \|_2 \left \| \x \right \|_2 \leq \frac{B}{\sqrt{d}} \left \| \bdelta^{(0)} \right \|_2.
\end{align}

\begin{align}
\label{Eq_local_lip_comp1}   
\left \| \frac{\partial  f(\x;\theta)}{\partial \W^{(\ell)}} \right \|_F &= \left \| \alpha \frac{\sigma_v}{\sqrt{n}} \frac{\sigma_w}{\sqrt{n}} \mathrm{diag} \left \{ \phi'(\g^{(\ell)}) \right \} \V^{(\ell)\top} \bdelta^{(\ell)} \x^{(\ell-1)\top} \right \|_F \\ \nonumber
&\leq \alpha C_\phi \sigma_v \sigma_w \frac{1}{\sqrt{n}} \| \V^{(\ell)\top} \bdelta^{(\ell)} \|_2 \frac{1}{\sqrt{n}} \| \x^{(\ell-1)} \|_2 \\ \nonumber
&\leq \alpha C_\phi \sigma_v \sigma_w \frac{1}{\sqrt{n}} \| \V^{(\ell)} \| \| \bdelta^{(\ell)} \|_2 \frac{1}{\sqrt{n}} \| \x^{(\ell-1)} \|_2 \\ \nonumber
&\leq 3 \alpha C_\phi \sigma_v \sigma_w \| \bdelta^{(\ell)} \|_2 \frac{1}{\sqrt{n}} \| \x^{(\ell-1)} \|_2. 
\end{align}

\begin{align}
\label{Eq_local_lip_comp2}   
\left \| \frac{\partial  f(\x;\theta)}{\partial \V^{(\ell)}} \right \|_F &= \left \| \alpha \frac{\sigma_v}{\sqrt{n}} \bdelta^{(\ell)} \phi(\g^{(\ell)})^\top \right \|_F = \left \| \alpha \frac{\sigma_v}{\sqrt{n}} \bdelta^{(\ell)} \phi( \frac{\sigma_w}{\sqrt{n}} \W^{(\ell)} \x^{(\ell-1)} )^\top  \right \|_F \\ \nonumber
& \leq \alpha C_\phi \sigma_v \sigma_w \| \bdelta^{(\ell)} \|_2 \frac{1}{\sqrt{n}} \| \W^{(\ell)} \| \frac{1}{\sqrt{n}} \| \x^{(\ell-1)} \|_2 \\ \nonumber
& \leq 3 \alpha C_\phi \sigma_v \sigma_w \| \bdelta^{(\ell)} \|_2 \frac{1}{\sqrt{n}} \| \x^{(\ell-1)} \|_2.
\end{align}

In Section~\ref{app:local_lip}.1 
we prove that $\frac{1}{\sqrt{n}} \| \x^{(\ell)} \|_2 \leq K_1$ and $\| \bdelta^{(\ell)} \|_2 \leq K_2$. Therefore,
\begin{align}
\label{Eq_local_lip_comp3}   
\| \J(\btheta) \|_F  \leq \sqrt{ |\mathcal{X}| \left ( (c_0 K_1)^2 + (c_1 K_2)^2 + L (c_2 K_1 K_2)^2 + L (c_3 K_1 K_2)^2  \right ) } = \tilde{K}. 
\end{align}

We turn to show that $\| \J(\btheta) - \J(\tilde{\btheta}) \|_F \leq K \| \btheta - \tilde{\btheta} \|_2$ for $\btheta,\tilde{\btheta} \in B(\btheta_0,C)$.
To simplify the notation, we add the tilde symbol above each term that depends on the $\tilde{\btheta}$  (rather than on $\btheta$), e.g., $\tilde{\x}^{(\ell)}$ denotes $\x^{(\ell)}(\x;\tilde{\btheta})$. 

\begin{align}
\label{Eq_local_lip_comp4}   
\| \J(\btheta) - \J(\tilde{\btheta}) \|_F^2  &= \sum \limits_{\x \in \mathcal{X}} \Bigg ( \left \| \frac{\partial  f(\x;\theta)}{\partial \w^{(L+1)}} - \frac{\partial  f(\x;\tilde{\btheta})}{\partial \tilde{\w}^{(L+1)}} \right \|_2^2 
\left \| \frac{\partial  f(\x;\theta)}{\partial \U} - \frac{\partial  f(\x;\tilde{\btheta})}{\partial \tilde{\U}} \right \|_F^2
\\ \nonumber
&\hspace{12mm} + \sum \limits_{\ell=1}^{L} \left \| \frac{\partial  f(\x;\theta)}{\partial \W^{(\ell)}} - \frac{\partial  f(\x;\tilde{\theta})}{\partial \tilde{\W}^{(\ell)}} \right \|_F^2 + \left \| \frac{\partial  f(\x;\theta)}{\partial \V^{(\ell)}} - \frac{\partial  f(\x;\tilde{\theta})}{\partial \tilde{\V}^{(\ell)}} \right \|_F^2  \Bigg ). 
\end{align}
Let us bound the terms in the sum.
\begin{align}
\label{Eq_local_lip_comp01_b}   
\left \| \frac{\partial  f(\x;\btheta)}{\partial \w^{(L+1)}} - \frac{\partial  f(\x;\tilde{\btheta})}{\partial \tilde{\w}^{(L+1)}} \right \|_2 = \frac{\sigma_w}{\sqrt{n}} \left \| \x^{(L)} - \tilde{\x}^{(L)} \right \|_2.
\end{align}

\begin{align}
\label{Eq_local_lip_comp01_b2}   
\left \| \frac{\partial  f(\x;\theta)}{\partial \U} - \frac{\partial  f(\x;\tilde{\btheta})}{\partial \tilde{\U}} \right \|_F = \frac{1}{\sqrt{d}} \left \| \bdelta^{(0)} \x^\top -  \tilde{\bdelta}^{(0)} \x^\top \right \|_F \leq \frac{B}{\sqrt{d}} \left \| \bdelta^{(0)} -  \tilde{\bdelta}^{(0)} \right \|_2.
\end{align}

\begin{align}
\label{Eq_local_lip_comp1_b}   
&\left \| \frac{\partial  f(\x;\theta)}{\partial \W^{(\ell)}} - \frac{\partial  f(\x;\tilde{\theta})}{\partial \tilde{\W}^{(\ell)}} \right \|_F \\ \nonumber
&= \left \| \underbrace{ \alpha \frac{\sigma_v \sigma_w}{\sqrt{n}} \mathrm{diag} \left \{ \phi'(\g^{(\ell)}) \right \} \V^{(\ell)\top} \bdelta^{(\ell)} }_{:= \bgamma^{(\ell)}} \frac{1}{\sqrt{n}} \x^{(\ell-1)\top}
- \underbrace{ \alpha \frac{\sigma_v \sigma_w}{\sqrt{n}} \mathrm{diag} \left \{ \phi'(\tilde{\g}^{(\ell)}) \right \} \tilde{\V}^{(\ell)\top} \tilde{\bdelta}^{(\ell)} }_{:= \tilde{\bgamma}^{(\ell)}} \frac{1}{\sqrt{n}} \tilde{\x}^{(\ell-1)\top}
\right \|_F \\ \nonumber
& \leq \left \| ( \bgamma^{(\ell)} - \tilde{\bgamma}^{(\ell)}) \frac{1}{\sqrt{n}} \x^{(\ell-1)\top} \right \|_F + \left \| \tilde{\bgamma}^{(\ell)} \frac{1}{\sqrt{n}} ( \x^{(\ell-1)\top} - \tilde{\x}^{(\ell-1)\top} ) \right \|_F \\ \nonumber
& \leq \frac{1}{\sqrt{n}} \left \| \x^{(\ell-1)} \right \|_2 \left \| \bgamma^{(\ell)} - \tilde{\bgamma}^{(\ell)} \right \|_2 +  \left \| \tilde{\bgamma}^{(\ell)} \right \|_2 \frac{1}{\sqrt{n}} \left \| \x^{(\ell-1)} - \tilde{\x}^{(\ell-1)}  \right  \|_2 \\ \nonumber
& \leq K_1 \left \| \bgamma^{(\ell)} - \tilde{\bgamma}^{(\ell)} \right \|_2 +  3 \alpha C_\phi \sigma_v \sigma_w K_2 \frac{1}{\sqrt{n}} \left \| \x^{(\ell-1)} - \tilde{\x}^{(\ell-1)}  \right  \|_2.
\end{align}

\begin{align}
\label{Eq_local_lip_comp2_b}   
&\left \| \frac{\partial  f(\x;\theta)}{\partial \V^{(\ell)}} - \frac{\partial  f(\x;\tilde{\theta})}{\partial \tilde{\V}^{(\ell)}} \right \|_F  \\ \nonumber
&= \left \|  \bdelta^{(\ell)} \frac{1}{\sqrt{n}} \underbrace{ \alpha \sigma_v \phi( \frac{\sigma_w}{\sqrt{n}} \W^{(\ell)} \x^{(\ell-1)} )^\top }_{:=\z^{(\ell)\top}}
- \tilde{\bdelta}^{(\ell)} \frac{1}{\sqrt{n}} \underbrace{ \alpha \sigma_v \phi( \frac{\sigma_w}{\sqrt{n}} \tilde{\W}^{(\ell)} \tilde{\x}^{(\ell-1)} )^\top }_{:=\tilde{\z}^{(\ell)\top}} \right \|_F \\ \nonumber
& \leq \left \| ( \bdelta^{(\ell)} - \tilde{\bdelta}^{(\ell)}) \frac{1}{\sqrt{n}} \z^{(\ell)\top} \right \|_F + \left \| \tilde{\bdelta}^{(\ell)} \frac{1}{\sqrt{n}} ( \z^{(\ell)\top} - \tilde{\z}^{(\ell)\top} ) \right \|_F \\ \nonumber
& \leq \frac{1}{\sqrt{n}} \left \| \z^{(\ell)} \right \|_2 \left \| \bdelta^{(\ell)} - \tilde{\bdelta}^{(\ell)} \right \|_2 +  \left \| \tilde{\bdelta}^{(\ell)} \right \|_2 \frac{1}{\sqrt{n}} \left \| \z^{(\ell)} - \tilde{\z}^{(\ell)}  \right  \|_2 \\ \nonumber
& \leq 3 \alpha C_\phi \sigma_v \sigma_w K_1 \left \| \bdelta^{(\ell)} - \tilde{\bdelta}^{(\ell)} \right \|_2 +  K_2 \frac{1}{\sqrt{n}} \left \| \z^{(\ell)} - \tilde{\z}^{(\ell)}  \right  \|_2.
\end{align}

Showing that 
$\left \| \bgamma^{(\ell)} - \tilde{\bgamma}^{(\ell)} \right \|_2, \frac{1}{\sqrt{n}} \left \| \x^{(\ell)} - \tilde{\x}^{(\ell)} \right  \|_2, \left \| \bdelta^{(\ell)} - \tilde{\bdelta}^{(\ell)} \right \|_2, \frac{1}{\sqrt{n}} \left \| \z^{(\ell)} - \tilde{\z}^{(\ell)} \right  \|_2 \leq \overline{K}\|\btheta - \tilde{\btheta}\|_2$ allows to obtain the required local Lipschitzness result for $\J(\btheta)$. 
However, as shown in 
Section~\ref{app:local_lip}.2, 
proving $\left \| \bdelta^{(\ell)} - \tilde{\bdelta}^{(\ell)} \right \|_2 \leq \overline{K}\|\btheta - \tilde{\btheta}\|_2$ requires that $\left \| \x^{(\ell)} - \tilde{\x}^{(\ell)} \right  \|_2 \leq \overline{K}\|\btheta - \tilde{\btheta}\|_2$ (without the $\frac{1}{\sqrt{n}}$ factor).

In Section~\ref{app:local_lip}.2 
we show that all the distances $\left \| \bgamma^{(\ell)} - \tilde{\bgamma}^{(\ell)} \right \|_2$, $\left \| \x^{(\ell)} - \tilde{\x}^{(\ell)} \right  \|_2$, $\left \| \bdelta^{(\ell)} - \tilde{\bdelta}^{(\ell)} \right \|_2$, $\left \| \z^{(\ell)} - \tilde{\z}^{(\ell)} \right  \|_2$ are indeed upper bounded by $\overline{K}\|\btheta - \tilde{\btheta}\|_2$.
Therefore, 
\begin{align}
\label{Eq_local_lip_comp3_b}   
\| \J(\btheta) - \J(\tilde{\btheta}) \|_F  \leq \sqrt{ |\mathcal{X}| \left ( (\overline{c_0} \overline{K})^2 + (\overline{c_1} \overline{K})^2 + L (\overline{c_2} \overline{K})^2 + L (\overline{c_3} \overline{K})^2  \right ) } \|\btheta - \tilde{\btheta}\|_2 = \tilde{\tilde{K}} \|\btheta - \tilde{\btheta}\|_2,
\end{align}
and the proof of Lemma~\ref{thm:local_lip} is finished with $K=\mathrm{max}(\tilde{K},\tilde{\tilde{K}})$.

\subsection{Auxiliary local boundness proofs}
\label{app:local_lip_aux1}

We prove by induction that $\frac{1}{\sqrt{n}} \| \x^{(\ell)} \|_2 \leq K_1$.

Base case: since $\|\x\|_2 \leq B$, we have with high probability over the random initialization of $\U \in \mathbb{R}^{n \times d}$ that $\frac{1}{\sqrt{n}} \| \x^{(0)} \|_2 = \frac{1}{\sqrt{n}} \| \frac{1}{\sqrt{d}} \U \x \|_2 \leq \frac{1}{\sqrt{d}} \frac{1}{\sqrt{n}} \| \U \| B \leq \frac{3}{\sqrt{d}}B$.

Assuming that $\frac{1}{\sqrt{n}} \| \x^{(\ell-1)} \|_2 \leq \tilde{K}_1$, we get 
\begin{align}
\label{Eq_local_lip_aux0}   
\frac{1}{\sqrt{n}} \| \x^{(\ell)} \|_2 &= \frac{1}{\sqrt{n}} \| \x^{(\ell-1)} + \alpha \frac{\sigma_v}{\sqrt{n}} \V^{(\ell)} \phi( \frac{\sigma_w}{\sqrt{n}} \W^{(\ell)} \x^{(\ell-1)} ) \|_2 \\ \nonumber
& \leq \left ( 1 + \alpha C_\phi \sigma_v \sigma_w \frac{1}{\sqrt{n}} \| \V^{(\ell)} \| \frac{1}{\sqrt{n}} \| \W^{(\ell)} \| \right ) \frac{1}{\sqrt{n}} \| \x^{(\ell-1)} \|_2 \\ \nonumber
& \leq \left ( 1 + 9 \alpha C_\phi \sigma_v \sigma_w \right ) \tilde{K}_1.
\end{align}
Therefore, we have that for all $\ell \in [L]: \frac{1}{\sqrt{n}} \| \x^{(\ell)} \|_2 \leq K_1 = ( 1 + 9 \alpha C_\phi \sigma_v \sigma_w)^L \frac{3}{\sqrt{d}}B$.

We prove by induction that $\| \bdelta^{(\ell)} \|_2 \leq K_2$. Recall that $\bdelta^{(\ell)} 
= \left ( \frac{\sigma_w}{\sqrt{n}}\w^{(L+1)\top} \frac{\partial \x^{(L)} }{\partial \x^{(L-1)}} \ldots   \frac{\partial \x^{(\ell+1)} }{\partial \x^{(\ell)}}  \right )^\top = \left ( \frac{\partial  \x^{(\ell+1)}}{\partial \x^{(\ell)}} \right )^\top \bdelta^{(\ell+1)}$.

Base case: $\| \bdelta^{(L)} \|_2 = \frac{\sigma_w}{\sqrt{n}} \| \w^{(L+1)} \|_2 \leq \frac{\sigma_w}{\sqrt{n}} 2\sqrt{n} = 2\sigma_w$.

Assuming that $\| \bdelta^{(\ell+1)} \|_2 \leq \tilde{K}_2$, we get
\begin{align}
\label{Eq_local_lip_aux1}   
\| \bdelta^{(\ell)} \|_2 &= \left \|  \left ( \I_n + \alpha  \frac{\sigma_w}{\sqrt{n}} \W^{(\ell+1)\top} \mathrm{diag} \left \{ \phi'(\g^{(\ell+1)}) \right \} \frac{\sigma_v}{\sqrt{n}} \V^{(\ell+1)\top} \right ) \bdelta^{(\ell+1)} \right \|_2  \\ \nonumber
&\leq \left ( 1 + \alpha C_\phi \sigma_v \sigma_w \frac{1}{\sqrt{n}} \| \V^{(\ell+1)} \| \frac{1}{\sqrt{n}} \| \W^{(\ell+1)} \| \right ) \| \bdelta^{(\ell+1)} \|_2 \\ \nonumber
& \leq \left ( 1 + 9 \alpha C_\phi \sigma_v \sigma_w \right ) \tilde{K}_2.
\end{align}
Therefore, we have that for all $\ell \in [L]: \| \bdelta^{(\ell)} \|_2 \leq K_2 = ( 1 + 9 \alpha C_\phi \sigma_v \sigma_w)^L 2\sigma_w$.

\subsection{Auxiliary local Lipschitzness proofs}
\label{app:local_lip_aux2}

Recall that $\btheta = \mathrm{vec}(\w^{(L+1)},\{\W^{(\ell)}\},\{\V^{(\ell)}\},\U)$. Therefore, we will repeatedly use $\| \W^{(\ell)} - \tilde{\W}^{(\ell)} \| \leq \| \W^{(\ell)} - \tilde{\W}^{(\ell)} \|_F \leq \| \btheta - \tilde{\btheta} \|_2$, and similarly for the other parameters. Also, for simplification we will use $\{c_i\}$ to denote constants that do not depend on $\btheta, \tilde{\btheta}, n$.

We prove by induction (together) that $\left \| \x^{(\ell)} - \tilde{\x}^{(\ell)} \right  \|_2 \leq \overline{K}\|\btheta - \tilde{\btheta}\|_2$ and also $\left \| \g^{(\ell)} - \tilde{\g}^{(\ell)} \right  \|_2 \leq \overline{K}\|\btheta - \tilde{\btheta}\|_2$.

Base case: $$\left \| \x^{(0)}(\x,\theta) - \x^{(0)}(\x,\tilde{\theta}) \right  \|_2 = \left \| \frac{1}{\sqrt{d}} \U \x - \frac{1}{\sqrt{d}} \tilde{\U} \x \right  \|_2 \leq \frac{1}{\sqrt{d}}  \| \U - \tilde{\U}  \| \|\x\|_2 \leq \frac{B}{\sqrt{d}} \| \btheta - \tilde{\btheta} \|_2,$$ 
and  
\begin{align}
\label{Eq_local_lip_aux3}  
\left \| \g^{(1)}(\x,\theta) - \g^{(1)}(\x,\tilde{\theta}) \right  \|_2 &= \left \| \frac{\sigma_w}{\sqrt{n}} \W^{(0)} \x^{(0)} - \frac{\sigma_w}{\sqrt{n}} \tilde{\W}^{(1)} \tilde{\x}^{(0)} \right  \|_2 \\ \nonumber
&\leq \left \| ( \W^{(0)}  - \tilde{\W}^{(1)} ) \frac{\sigma_w}{\sqrt{n}} \x^{(0)} \right  \|_2 + \left \| \frac{\sigma_w}{\sqrt{n}} \tilde{\W}^{(1)} ( \x^{(0)} - \tilde{\x}^{(0)}) \right  \|_2 \\ \nonumber
&\leq  \| \W^{(0)}  - \tilde{\W}^{(1)} \| \frac{\sigma_w}{\sqrt{n}} \| \x^{(0)} \|_2 +  \frac{\sigma_w}{\sqrt{n}} \| \tilde{\W}^{(1)} \| \| \x^{(0)} - \tilde{\x}^{(0)} \|_2  \\ \nonumber
&\leq  \sigma_w K_1 \|\btheta - \tilde{\btheta}\|_2 +  3\sigma_w \frac{B}{\sqrt{d}} \| \btheta - \tilde{\btheta} \|_2.
\end{align}
Thus, $\left \| \x^{(0)}(\x,\theta) - \x^{(0)}(\x,\tilde{\theta}) \right  \|_2, \left \| \g^{(1)}(\x,\theta) - \g^{(1)}(\x,\tilde{\theta}) \right  \|_2 \leq c_1 \| \btheta - \tilde{\btheta} \|_2$.
 
Assuming that $\| \x^{(\ell-1)} - \tilde{\x}^{(\ell-1)}  \|_2, \| \g^{(\ell)} - \tilde{\g}^{(\ell)}  \|_2 \leq \tilde{K}\|\btheta - \tilde{\btheta}\|_2$, we get
\begin{align}
\label{Eq_local_lip_aux4}   
\| \x^{(\ell)} - \tilde{\x}^{(\ell)}  \|_2 &= \left \|  \x^{(\ell-1)} + \alpha \frac{\sigma_v}{\sqrt{n}} \V^{(\ell)} \phi(\g^{(\ell)}) - \tilde{\x}^{(\ell-1)} - \alpha \frac{\sigma_v}{\sqrt{n}} \tilde{\V}^{(\ell)} \phi(\tilde{\g}^{(\ell)}) \right \|_2  \\ \nonumber
&\leq \| \x^{(\ell-1)} - \tilde{\x}^{(\ell-1)} \|_2 + \alpha \left \| \frac{\sigma_v}{\sqrt{n}} \V^{(\ell)} \phi(\g^{(\ell)}) - \frac{\sigma_v}{\sqrt{n}} \tilde{\V}^{(\ell)} \phi(\tilde{\g}^{(\ell)}) \right \|_2 \\ \nonumber
& \leq \tilde{K}\|\btheta - \tilde{\btheta}\|_2 + \alpha \| \V^{(\ell)}  - \tilde{\V}^{(\ell)} \| \frac{\sigma_v}{\sqrt{n}} \| \phi(\g^{(\ell)}) \|_2 +  \frac{\sigma_v}{\sqrt{n}} \| \tilde{\V}^{(\ell)} \| \| \phi(\g^{(\ell)}) - \phi(\tilde{\g}^{(\ell)}) \|_2   \\ \nonumber
& \leq \tilde{K}\|\btheta - \tilde{\btheta}\|_2 + \alpha C_\phi \frac{\sigma_v}{\sqrt{n}} \| \frac{\sigma_w}{\sqrt{n}} \W^{(\ell)} \x^{(\ell-1)} \|_2 \|\btheta - \tilde{\btheta}\|_2 +  3\sigma_v C_\phi \| \g^{(\ell)} - \tilde{\g}^{(\ell)} \|_2   \\ \nonumber
& \leq (\tilde{K} + 3\sigma_v C_\phi) \|\btheta - \tilde{\btheta}\|_2 + 3 \alpha C_\phi \sigma_v \sigma_w \frac{1}{\sqrt{n}} \| \x^{(\ell-1)} \|_2 \|\btheta - \tilde{\btheta}\|_2   \\ \nonumber
& \leq (\tilde{K} + 3\sigma_v C_\phi + 3 \alpha C_\phi \sigma_v \sigma_w K_1)  \|\btheta - \tilde{\btheta}\|_2 \leq c_2 \|\btheta - \tilde{\btheta}\|_2.
\end{align}
\begin{align}
\label{Eq_local_lip_aux5}   
\left \| \g^{(\ell+1)} - \tilde{\g}^{(\ell+1)} \right  \|_2 &= \left \| \frac{\sigma_w}{\sqrt{n}} \W^{(\ell+1)} \x^{(\ell)} - \frac{\sigma_w}{\sqrt{n}} \tilde{\W}^{(\ell+1)} \tilde{\x}^{(\ell)} \right  \|_2 \\ \nonumber
&\leq  \| \W^{(\ell+1)}  - \tilde{\W}^{(\ell+1)} \| \frac{\sigma_w}{\sqrt{n}} \| \x^{(\ell)} \|_2 +  \frac{\sigma_w}{\sqrt{n}} \| \tilde{\W}^{(\ell+1)} \| \| \x^{(\ell)} - \tilde{\x}^{(\ell)} \|_2  \\ \nonumber
&\leq  \sigma_w K_1 \|\btheta - \tilde{\btheta}\|_2 +  3\sigma_w c_2 \| \btheta - \tilde{\btheta} \|_2 \leq c_3 \|\btheta - \tilde{\btheta}\|_2.
\end{align}
Therefore, we have that for all $\ell \in [L]: \| \x^{(\ell)} - \tilde{\x}^{(\ell)}  \|_2, \| \g^{(\ell)} - \tilde{\g}^{(\ell)}  \|_2 \leq c_3^L \|\btheta - \tilde{\btheta}\|_2 \leq \overline{K}\|\btheta - \tilde{\btheta}\|_2$.

The proof for $\left \| \z^{(\ell)} - \tilde{\z}^{(\ell)} \right  \|_2 \leq \overline{K}\|\btheta - \tilde{\btheta}\|_2$ follows from directly from $\| \g^{(\ell)} - \tilde{\g}^{(\ell)}  \|_2 \leq c_3^L \|\btheta - \tilde{\btheta}\|_2$:
\begin{align}
\label{Eq_local_lip_aux6}   
\left \| \z^{(\ell)} - \tilde{\z}^{(\ell)} \right  \|_2 &= \left \| \alpha \sigma_v \phi( \g^{(\ell)} ) - \alpha \sigma_v \phi( \tilde{\g}^{(\ell)} ) \right  \|_2 \\ \nonumber
&\leq  \alpha C_\phi \sigma_v \left \| \g^{(\ell)} - \tilde{\g}^{(\ell)} \right  \|_2 \leq \alpha C_\phi \sigma_v c_3^L \|\btheta - \tilde{\btheta}\|_2 \leq \overline{K}\|\btheta - \tilde{\btheta}\|_2.
\end{align}

We turn to prove by induction that $\left \| \bdelta^{(\ell)} - \tilde{\bdelta}^{(\ell)} \right  \|_2 \leq \overline{K}\|\btheta - \tilde{\btheta}\|_2$.

Base case: $\left \| \bdelta^{(L)} - \tilde{\bdelta}^{(L)} \right  \|_2 = \left \| \frac{\sigma_w}{\sqrt{n}}\w^{(L+1)} - \frac{\sigma_w}{\sqrt{n}}\tilde{\w}^{(L+1)} \right  \|_2 \leq \frac{\sigma_w}{\sqrt{n}} \| \btheta - \tilde{\btheta} \|_2 \leq \tilde{K} \| \btheta - \tilde{\btheta} \|_2$.

Assuming that $\| \bdelta^{(\ell+1)} - \tilde{\bdelta}^{(\ell+1)}  \|_2 \leq \tilde{K}\|\btheta - \tilde{\btheta}\|_2$, we get
\begin{align}
\label{Eq_local_lip_aux7}   
&\| \bdelta^{(\ell)} - \tilde{\bdelta}^{(\ell)}  \|_2  =\Bigg \|  \left ( \I_n + \alpha \frac{\sigma_v}{\sqrt{n}} \V^{(\ell+1)} \mathrm{diag} \left \{ \phi'(\g^{(\ell+1)}) \right \} \frac{\sigma_w}{\sqrt{n}} \W^{(\ell+1)} \right )^\top \bdelta^{(\ell+1)}  \\ \nonumber
&\hspace{25mm}-  \left ( \I_n + \alpha \frac{\sigma_v}{\sqrt{n}} \tilde{\V}^{(\ell+1)} \mathrm{diag} \left \{ \phi'(\tilde{\g}^{(\ell+1)}) \right \} \frac{\sigma_w}{\sqrt{n}} \tilde{\W}^{(\ell+1)} \right )^\top \tilde{\bdelta}^{(\ell+1)} \Bigg \|_2  \\ \nonumber
&\leq \| \bdelta^{(\ell+1)} - \tilde{\bdelta}^{(\ell+1)} \|_2  + \left \| \left ( \alpha \frac{\sigma_v}{\sqrt{n}} \V^{(\ell+1)} \mathrm{diag} \left \{ \phi'(\g^{(\ell+1)}) \right \} \frac{\sigma_w}{\sqrt{n}} \W^{(\ell+1)} \right )^\top ( \bdelta^{(\ell+1)} - \tilde{\bdelta}^{(\ell+1)}) \right \|_2  \\ \nonumber
& \hspace{5mm} + \left \| \left ( \alpha \frac{\sigma_v}{\sqrt{n}} \V^{(\ell+1)} \mathrm{diag} \left \{ \phi'(\g^{(\ell+1)}) \right \} \frac{\sigma_w}{\sqrt{n}} \W^{(\ell+1)} - \alpha \frac{\sigma_v}{\sqrt{n}} \tilde{\V}^{(\ell+1)} \mathrm{diag} \left \{ \phi'(\tilde{\g}^{(\ell+1)}) \right \} \frac{\sigma_w}{\sqrt{n}} \tilde{\W}^{(\ell+1)} \right )^\top \tilde{\bdelta}^{(\ell+1)})  \right \|_2   \\ \nonumber
& \leq ( \tilde{K} + 9 \alpha \sigma_v \sigma_w C_\phi \tilde{K}  ) \|\btheta - \tilde{\btheta}\|_2  \\ \nonumber
& \hspace{5mm} + K_2 \left \| \alpha \frac{\sigma_v}{\sqrt{n}} \V^{(\ell+1)} \mathrm{diag} \left \{ \phi'(\g^{(\ell+1)}) \right \} \frac{\sigma_w}{\sqrt{n}} \W^{(\ell+1)} - \alpha \frac{\sigma_v}{\sqrt{n}} \tilde{\V}^{(\ell+1)} \mathrm{diag} \left \{ \phi'(\tilde{\g}^{(\ell+1)}) \right \} \frac{\sigma_w}{\sqrt{n}} \tilde{\W}^{(\ell+1)}   \right \|   \\ \nonumber
& \leq c_4 \|\btheta - \tilde{\btheta}\|_2 + \alpha K_2 \left \| \frac{\sigma_v}{\sqrt{n}} \V^{(\ell+1)} \mathrm{diag} \left \{ \phi'(\g^{(\ell+1)}) \right \} \frac{\sigma_w}{\sqrt{n}} ( \W^{(\ell+1)} - \tilde{\W}^{(\ell+1)} )  \right \|   \\ \nonumber
& \hspace{5mm} + \alpha K_2 \left \| \left ( \frac{\sigma_v}{\sqrt{n}} \V^{(\ell+1)} \mathrm{diag} \left \{ \phi'(\g^{(\ell+1)}) \right \} - \frac{\sigma_v}{\sqrt{n}} \tilde{\V}^{(\ell+1)} \mathrm{diag} \left \{ \phi'(\tilde{\g}^{(\ell+1)}) \right \}  \right )  \frac{\sigma_w}{\sqrt{n}} \tilde{\W}^{(\ell+1)}  \right \|
\end{align}
\begin{align}
\label{Eq_local_lip_aux8}   
& \leq (c_4 + \alpha K_2 3 \sigma_v  \frac{\sigma_w}{\sqrt{n}} C_\phi ) \|\btheta - \tilde{\btheta}\|_2   \\ \nonumber
& \hspace{5mm} + c_5   
\left (  \left \| \frac{\sigma_v}{\sqrt{n}} ( \V^{(\ell+1)} - \tilde{\V}^{(\ell+1)} ) \mathrm{diag} \left \{ \phi'(\tilde{\g}^{(\ell+1)}) \right \}   \right \|
+ \left \|  \frac{\sigma_v}{\sqrt{n}} \V^{(\ell+1)} ( \mathrm{diag} \left \{ \phi'(\g^{(\ell+1)}) \right \} - \mathrm{diag} \left \{ \phi'(\tilde{\g}^{(\ell+1)}) \right \} ) \right \|   \right ) \\ \nonumber
& \leq ( c_6 + c_5 \frac{\sigma_v}{\sqrt{n}} C_\phi ) \|\btheta - \tilde{\btheta}\|_2 + 3 c_5 \sigma_v C_\phi \left \|  \g^{(\ell+1)} - \tilde{\g}^{(\ell+1)} \right  \|_2  \\ \nonumber
& \leq ( c_6 + c_5 \frac{\sigma_v}{\sqrt{n}} C_\phi ) \|\btheta - \tilde{\btheta}\|_2 + 3 c_5 \sigma_v C_\phi c_3^L \|\btheta - \tilde{\btheta}\|_2  \leq c_7 \|\btheta - \tilde{\btheta}\|_2.
\end{align}
Therefore, we have that for all $\ell \in [L]: \| \bdelta^{(\ell)} - \tilde{\bdelta}^{(\ell)}  \|_2 \leq c_7^L \|\btheta - \tilde{\btheta}\|_2 \leq \overline{K}\|\btheta - \tilde{\btheta}\|_2$.

It is left to prove that $\left \| \bgamma^{(\ell)} - \tilde{\bgamma}^{(\ell)} \right \|_2 \leq \overline{K}\|\btheta - \tilde{\btheta}\|_2$. This is achieved by the previous results for $\| \bdelta^{(\ell)} - \tilde{\bdelta}^{(\ell)}  \|_2$ and $\| \g^{(\ell)} - \tilde{\g}^{(\ell)}  \|_2$:
\begin{align}
\label{Eq_local_lip_aux9}   
&\left \| \bgamma^{(\ell)} - \tilde{\bgamma}^{(\ell)} \right \|_2  = 
\left \| \alpha \frac{\sigma_v \sigma_w}{\sqrt{n}} \mathrm{diag} \left \{ \phi'(\g^{(\ell)}) \right \} \V^{(\ell)\top} \bdelta^{(\ell)} - 
\alpha \frac{\sigma_v \sigma_w}{\sqrt{n}} \mathrm{diag} \left \{ \phi'(\tilde{\g}^{(\ell)}) \right \} \tilde{\V}^{(\ell)\top} \tilde{\bdelta}^{(\ell)} \right \|_2  \\ \nonumber
& \leq \alpha \sigma_w  \left \| \frac{\sigma_v}{\sqrt{n}} \mathrm{diag} \left \{ \phi'(\g^{(\ell)}) \right \} \V^{(\ell)\top} ( \bdelta^{(\ell)} - \tilde{\bdelta}^{(\ell)} ) \right \|_2 \\ \nonumber
&\hspace{5mm} + \alpha \sigma_w \left \| \frac{\sigma_v}{\sqrt{n}} \left ( \mathrm{diag} \left \{ \phi'(\g^{(\ell)}) \right \} \V^{(\ell)\top} - \mathrm{diag} \left \{ \phi'(\tilde{\g}^{(\ell)}) \right \} \tilde{\V}^{(\ell)\top} \right ) \tilde{\bdelta}^{(\ell)} \right \|_2 \\ \nonumber
&\leq 3 \alpha \sigma_w \sigma_v C_\phi \| \bdelta^{(\ell)} - \tilde{\bdelta}^{(\ell)}  \|_2  \\ \nonumber
& \hspace{5mm} + \alpha \sigma_w K_2 \left (  \left \| \frac{\sigma_v}{\sqrt{n}} ( \V^{(\ell)} - \tilde{\V}^{(\ell)}  ) \mathrm{diag} \left \{ \phi'(\g^{(\ell)}) \right \}  \right  \|
+ \left  \|  \frac{\sigma_v}{\sqrt{n}} \tilde{\V}^{(\ell)} ( \mathrm{diag} \left \{ \phi'(\g^{(\ell)}) \right \} - \mathrm{diag} \left \{ \phi'(\tilde{\g}^{(\ell)}) \right \}  )  \right  \| 
 \right )  \\ \nonumber
& \leq ( c_8 c_7^L + c_9 \frac{\sigma_v}{\sqrt{n}} C_\phi ) \|\btheta - \tilde{\btheta}\|_2 + 3 c_9 \sigma_v C_\phi \left \|  \g^{(\ell+1)} - \tilde{\g}^{(\ell+1)} \right  \|_2  \\ \nonumber
&\leq c_{10} \|\btheta - \tilde{\btheta}\|_2  \leq \overline{K} \|\btheta - \tilde{\btheta}\|_2.
\end{align}

To conclude, we showed that for $\btheta,\tilde{\btheta} \in B(\btheta_0,C)$ there exists $\overline{K}>0$ (that does not depend on $\btheta, \tilde{\btheta}, n$) such that all the distances $\left \| \bgamma^{(\ell)} - \tilde{\bgamma}^{(\ell)} \right \|_2$, $\left \| \x^{(\ell)} - \tilde{\x}^{(\ell)} \right  \|_2$, $\left \| \bdelta^{(\ell)} - \tilde{\bdelta}^{(\ell)} \right \|_2$, $\left \| \z^{(\ell)} - \tilde{\z}^{(\ell)} \right  \|_2$ are upper bounded by $\overline{K}\|\btheta - \tilde{\btheta}\|_2$.

\newpage

\section{Additional Empirical \tomtb{NTK} Results}
\label{app:additional_exp}


\subsection{\tomtb{Functions with Scalar Inputs}}

In this section we provide more experiments and details on the experimental setting that are missing in the main body of the paper, due to space limitation. 

First, let us state the underlying ground truth function whose samples are used in the interpolation experiments
\begin{align}
\label{Eq_app_gt_func}   
f(\beta) = \frac{1}{2}\mathrm{cos}(\beta) + \mathrm{sin}(4\beta), \hspace{5mm} -\pi \leq \beta \leq \pi.
\end{align}
We treat the samples $\{ \beta_i \}$ as points on the sphere (circle) in $\mathbb{R}^2$, i.e., samples of $\{\x\in\mathbb{R}^2 : x_1^2+x_2^2 = 1\}$, since any point on the sphere has 1-to-1 mapping to an angle $\beta$, and vice versa ($\beta \xrightarrow{} (\mathrm{cos}\beta, \mathrm{sin}\beta)$). This is motivated by the proof in \citep{jacot2018neural} that restricting the NTK to the unit sphere yields $\lambda_{min}(\bTheta)>0$, which is required for the NTK theory.
Note that we do not examine or compare reconstruction errors in this paper, and $f(\beta)$ is given here for reproducibility reasons. 

\begin{figure}[h]
  \centering
  \subfigure[{\scriptsize NTKs (normalized to unit peak) $L=5$}]{%
    \label{fig:NTK_mlp_resnet_L5_kernels_appendix}
    \includegraphics[width=140pt]{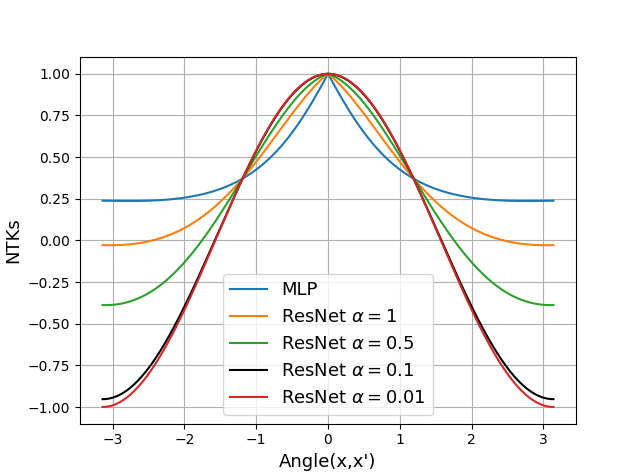}
  }
  \subfigure[{\scriptsize Interpolation with 6 samples $L=5$}]{%
    \label{fig:NTK_mlp_resnet_L5_regression_N6_appendix}
    \includegraphics[width=140pt]{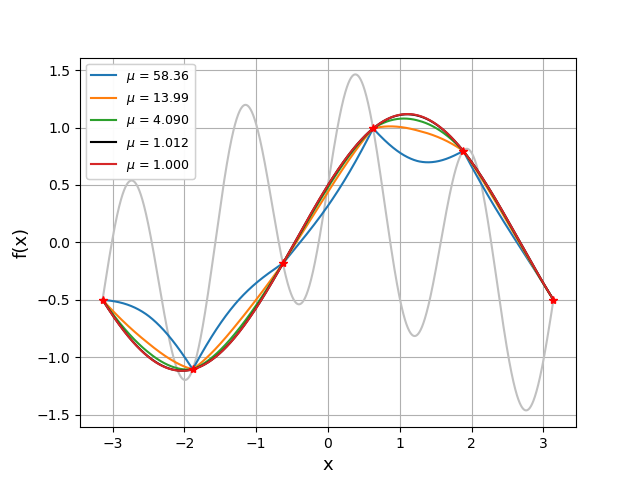}
  }
  \subfigure[{\scriptsize Interpolation with 10 samples $L=5$}]{%
    \label{fig:NTK_mlp_resnet_L5_regression_N10_appendix}
    \includegraphics[width=140pt]{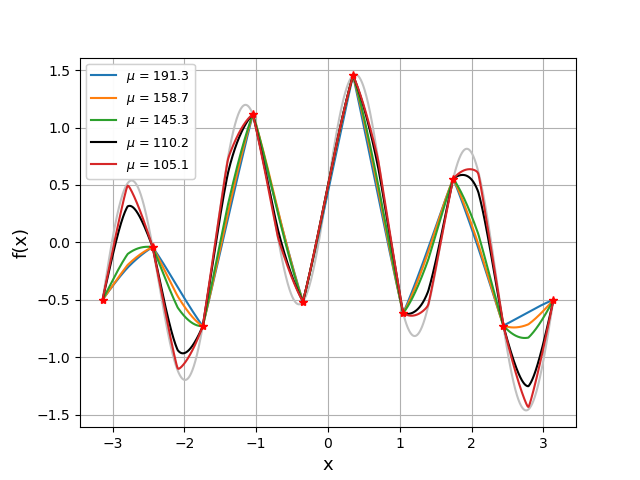}
  }


  \subfigure[{\scriptsize NTKs (normalized to unit peak) $L=15$}]{%
    \label{fig:NTK_mlp_resnet_L15_kernels_appendix}
    \includegraphics[width=140pt]{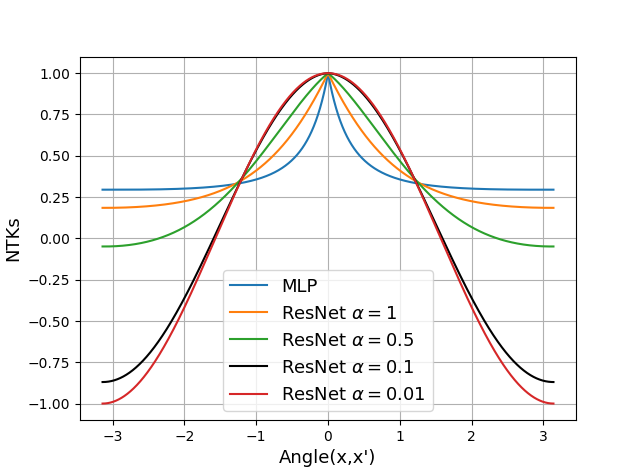}
  }
  \subfigure[{\scriptsize Interpolation with 6 samples $L=15$}]{%
    \label{fig:NTK_mlp_resnet_L15_regression_N6_appendix}
    \includegraphics[width=140pt]{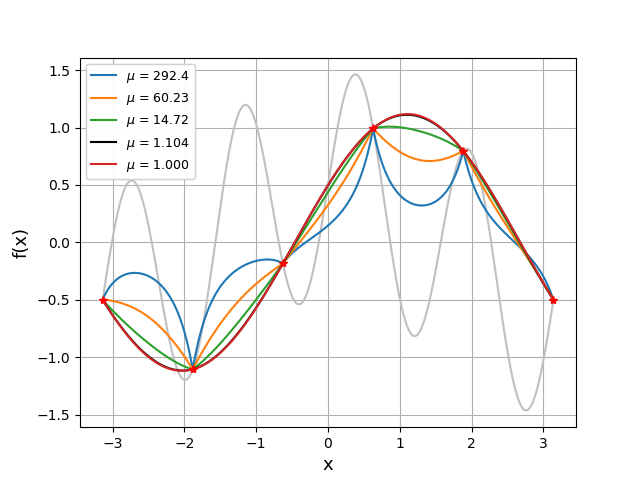}
  }
  \subfigure[{\scriptsize Interpolation with 10 samples $L=15$}]{%
    \label{fig:NTK_mlp_resnet_L15_regression_N10_appendix}
    \includegraphics[width=140pt]{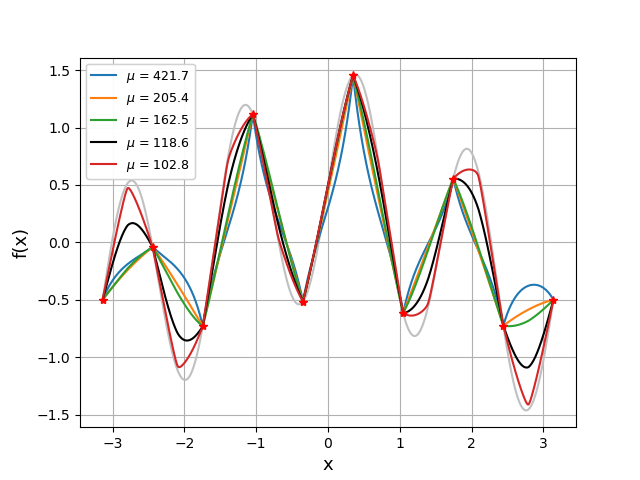}
  }

    \caption{NTKs for MLP and ResNet (for different values of $\alpha$) with $L=5$ (top) and $L=15$ (bottom) nonlinear layers, ReLU nonlinearities, $\sigma_v=\sigma_w=1$, for inputs on the sphere (circle) in $\mathbb{R}^2$.
    (\ref{fig:NTK_mlp_resnet_L5_kernels_appendix}),(\ref{fig:NTK_mlp_resnet_L15_kernels_appendix}): The kernels shape.  
    (\ref{fig:NTK_mlp_resnet_L5_regression_N6_appendix})-(\ref{fig:NTK_mlp_resnet_L5_regression_N10_appendix}), (\ref{fig:NTK_mlp_resnet_L15_regression_N6_appendix})-(\ref{fig:NTK_mlp_resnet_L15_regression_N10_appendix}): Interpolations by the closed-form solutions, measured by $\mu(\cdot)$ defined in \eqref{Eq_L2_der2_meas2}. 
    Note that the legend in (\ref{fig:NTK_mlp_resnet_L5_kernels_appendix}),(\ref{fig:NTK_mlp_resnet_L15_kernels_appendix}) applies to all the figures. 
    }
\label{fig:NTK_mlp_resnet_L15_appendix}     
\end{figure}

Next, we present in Figure~\ref{fig:NTK_mlp_resnet_L15_appendix} an extended version of Figure~\ref{fig:NTK_mlp_resnet_L15} that includes also the results of ResNet NTK for an extremely small value of $\alpha$, namely $\alpha=0.01$. 
It can be seen that this small $\alpha$ does not significantly affect the kernel shape and interpolations' smoothness compared to the moderate $\alpha=0.1$.

\begin{figure}
  \centering
  \subfigure[{\scriptsize Kernels (normalized to unit peak)}]{%
    \label{fig:NTK_mlp_resnet_L7_appendix_kernels}
    \includegraphics[width=140pt]{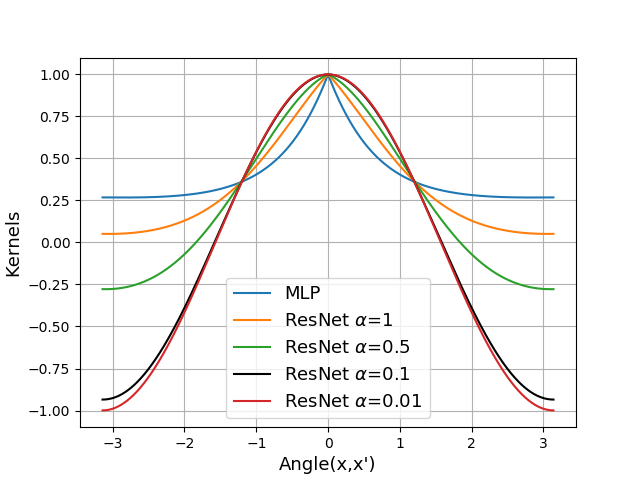}
  }
  \subfigure[{\scriptsize Interpolation with 15 random samples}]{%
    \label{fig:NTK_mlp_resnet_L7_appendix_regression_N6}
    \includegraphics[width=140pt]{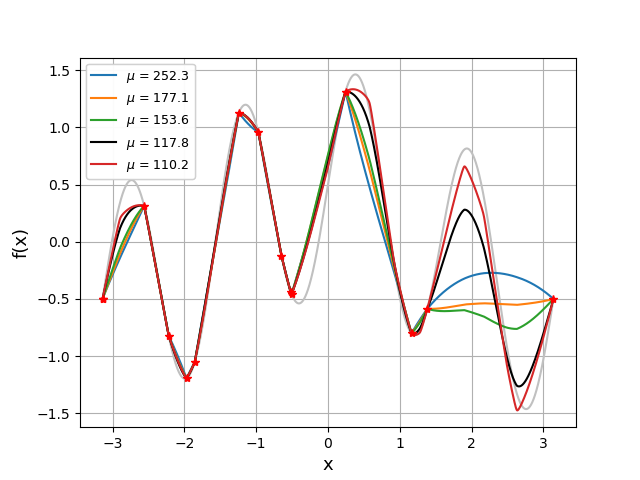}
  }

    \caption{NTKs for MLP and ResNet (for different values of $\alpha$) with $L=7$ nonlinear layers, ReLU nonlinearities, $\sigma_v=\sigma_w=1$, for inputs on the sphere (circle) in $\mathbb{R}^2$.
    (\ref{fig:NTK_mlp_resnet_L7_appendix_kernels}): The kernels shape.  
    (\ref{fig:NTK_mlp_resnet_L7_appendix_regression_N6}): Interpolations by the closed-form solutions, measured by $\mu(\cdot)$ defined in \eqref{Eq_L2_der2_meas2}. 
    Note that the legend in (\ref{fig:NTK_mlp_resnet_L7_appendix_kernels}) applies to all the figures. 
    }
\label{fig:NTK_mlp_resnet_L7_appendix}     
\end{figure}

More NTK results, this time for $L=7$ nonlinear layers and 15 random samples, are presented in Figure~\ref{fig:NTK_mlp_resnet_L7_appendix}.

\begin{figure}
    \centering\includegraphics[width=170pt]{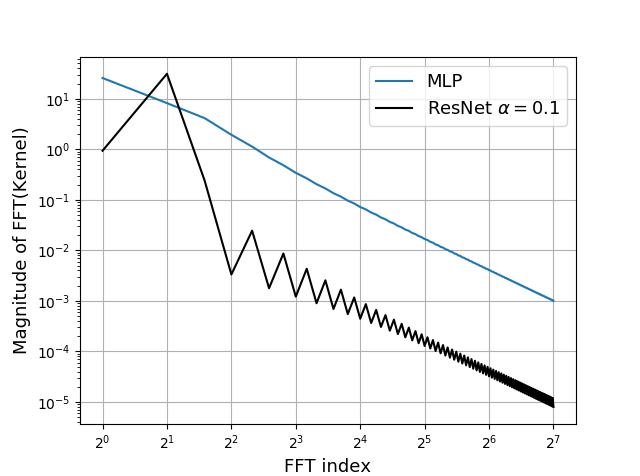}
    \caption{Magnitude of the first 128 FFT elements (out of 4096) of the NTKs for MLP and ResNet (with $\alpha=0.1$) with $L=5$ nonlinear layers, ReLU nonlinearities, $\sigma_v=\sigma_w=1$, for inputs on the sphere (circle) in $\mathbb{R}^2$.
    }
\label{fig:Asysp_NTKs_mlp_resnet_L5_FFT}     
\end{figure}

Finally, in Figure~\ref{fig:Asysp_NTKs_mlp_resnet_L5_FFT} we present, in logarithmic scale, the FFT spectrums of the NTKs of ResNet with $\alpha=0.1$ and MLP, both with $L=5$ nonlinear layers. 
This figure demonstrates our claim below \eqref{Eq_L2_der2_meas2}: 
Since the decay rates of the FFT coefficients of the different kernels are approximately different only by a factor, we find the measure $\mu(f)$, {\em which depends also on the resulted interpolation and not only on the kernel}, to be more informative than comparing the FFT of the kernels. For example, multiplying the ResNet NTK by a large constant factor will make the magnitude of its FFT larger than the magnitude of the FFT of MLP NTK. Yet, it will not change the results of the kernel regression.

\subsection{\tomtb{Functions with Multidimensional Inputs}}
\label{app:additional_exp_multidimensional}

\tomtb{
In this section we briefly demonstrate the smoothness distinction between ResNet and MLP NTKs for multidimensional samples.}

\tomtb{
First, we extend the previous one-dimensional experiments to two dimensions.
We modify the underlying ground truth function in \eqref{Eq_app_gt_func}  to
\begin{align}
\label{Eq_app_gt_func_2D}   
f(\beta,\xi) = \left (\frac{1}{2}\mathrm{cos}(\beta) + \mathrm{sin}(4\beta) \right )\mathrm{sin}(\xi), \hspace{5mm} -\pi \leq \beta \leq \pi, \hspace{2mm} 0 \leq \xi \leq \pi.
\end{align}
Here, the samples $\{ (\beta_i,\xi_i) \}$ can be treated as points on the sphere in $\mathbb{R}^3$, i.e., samples of $\{\x\in\mathbb{R}^3 : \|\x\|_2 = 1\}$, since any point on the sphere has 1-to-1 mapping to a pair of angles $(\beta,\xi)$, and vice versa ($(\beta,\xi) \xrightarrow{} (\mathrm{cos}\beta \mathrm{sin}\xi, \mathrm{sin}\beta \mathrm{sin}\xi, \mathrm{cos}\xi)$). Recall that for unique samples (i.e., $\x_i \neq \x_j$ for $i \neq j$) from the unit sphere we get $\lambda_{min}(\bTheta)>0$, which is required for the NTK theory \citep{jacot2018neural}.
}

\begin{figure}
  \centering
    \includegraphics[width=210pt]{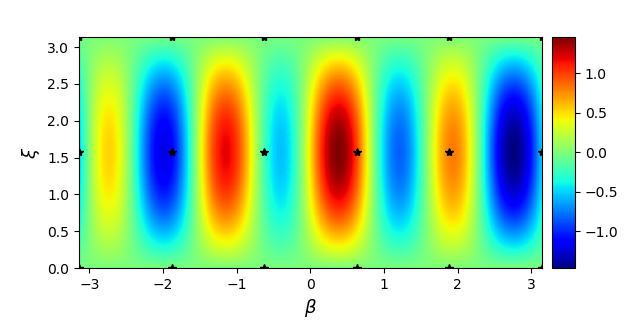}
    \includegraphics[width=210pt]{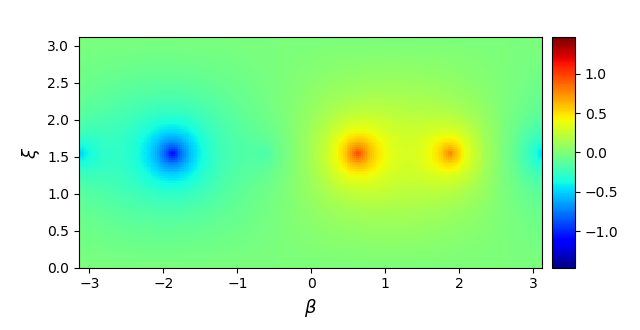}
    \includegraphics[width=210pt]{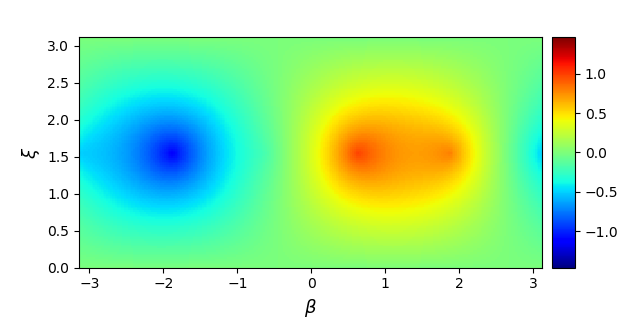}
    \includegraphics[width=210pt]{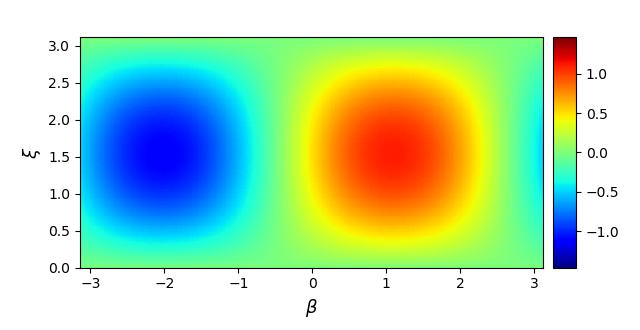}
  
    \caption{Two-dimensional interpolations from $6 \times 3$ samples using
 MLP and ResNet NTKs with $L=15$ nonlinear layers, ReLU nonlinearities, $\sigma_v=\sigma_w=1$, for inputs on the sphere in $\mathbb{R}^3$. From left to right and top to bottom: the underlying function with $6 \times 3$ samples, interpolation by the MLP NTK ($\mu=1.47\mathrm{e}4$), interpolation by the ResNet NTK with $\alpha=1$ ($\mu=2.33\mathrm{e}3$), and interpolation by the ResNet NTK with $\alpha=0.1$ ($\mu=1.55\mathrm{e}1$).  
    }
\label{fig:NTK_mlp_resnet_2D_N6x3_appendix}     
\end{figure}

\begin{figure}
  \centering
    \includegraphics[width=210pt]{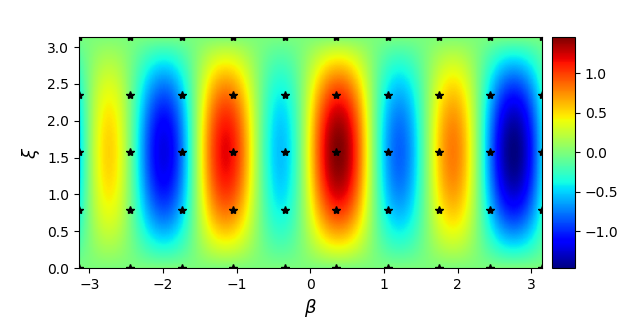}
    \includegraphics[width=210pt]{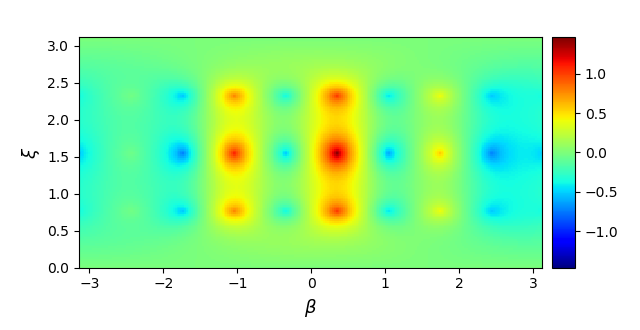}
    \includegraphics[width=210pt]{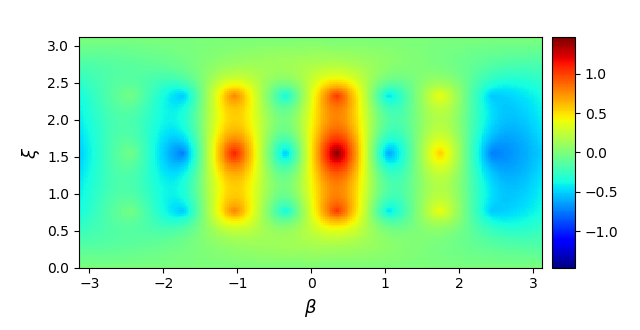}
    \includegraphics[width=210pt]{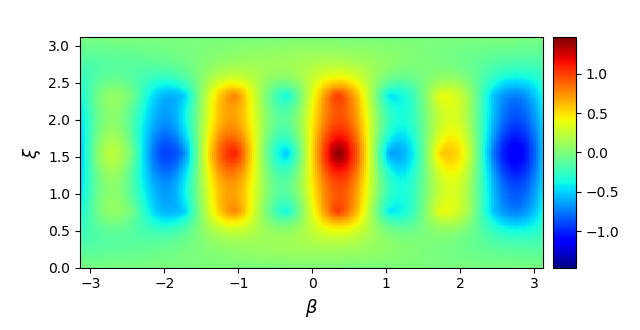}
  
    \caption{Two-dimensional interpolations from $10 \times 5$ samples using
 MLP and ResNet NTKs with $L=15$ nonlinear layers, ReLU nonlinearities, $\sigma_v=\sigma_w=1$, for inputs on the sphere in $\mathbb{R}^3$. From left to right and top to bottom: the underlying function with $10 \times 5$ samples, interpolation by the MLP NTK ($\mu=2.59\mathrm{e}4$), interpolation by the ResNet NTK with $\alpha=1$ ($\mu=1.24\mathrm{e}4$), and interpolation by the ResNet NTK with $\alpha=0.1$ ($\mu=7.18\mathrm{e}3$).  
    }
\label{fig:NTK_mlp_resnet_2D_N10x5_appendix}     
\end{figure}

\tomtb{
Figures~\ref{fig:NTK_mlp_resnet_2D_N6x3_appendix} and \ref{fig:NTK_mlp_resnet_2D_N10x5_appendix}  show the interpolation results of the ResNet NTK with different values of $\alpha$ and the MLP NTK, for $6 \times 3$ uniform samples and for $10 \times 5$ uniform samples. We also report there the measure $\mu(\cdot)$ defined in \eqref{Eq_L2_der2_meas2} (straightforwardly extended to the 2D case).
Similar to the one-dimensional case, it can be seen that the interpolation results of ResNet NTK are smoother, especially with small $\alpha$.}

\tomtb{
We turn to examine the NTKs for high-dimensional data; 
specifically, the MNIST dataset, 
which includes images of size $28 \times 28$ of handwritten digits. 
As visualizing and measuring smoothness is difficult for high-dimensional data, we consider binary classification tasks, in which we interpolate (``overfit'') the training set and report accuracy results on the test set. This links the different smoothness of the different NTKs with their generalization.} 

\tomtb{
We consider the digits ``0'' and ``8'' and label them with $y \in \{+1,-1\}$. The training set includes $N/2$ samples from each class, where $N \in \{50,100\}$, and the test set includes 1000 images from each class.
The samples are presented as vectors in $\mathbb{R}^{28^2}$. 
The mean vector of the training set is subtracted from all samples and each sample is normalized to have unit Euclidean norm. 
We examine the MLP NTK and the ResNet NTK with $\alpha \in \{1, 0.1\}$ and $L \in \{ 5,15,30 \}$. The classification of a test sample is made according to the sign of the $\ell_2$ kernel regression (using the closed-form expression in \eqref{Eq_kernel_sol}).
The results are presented in Table~\ref{table:mnist_class}. They demonstrate the advantage of using a smoother ResNet NTK.}

\begin{table}[h]
\small
\renewcommand{\arraystretch}{1.3}
\caption{Accuracy results of different NTKs for binary MNIST classification tasks.} \label{table:mnist_class}
\centering
\tomtb{
    \begin{tabular}{ | l | l | l | l |}
    \hline
    50 training samples  &  MLP NTK & ResNet NTK $\alpha=1$ & ResNet NTK $\alpha=0.1$ \\ \hline
  $L=5$  & 0.9625 & 0.967 & 0.972 \\ \hline
$L=15$  & 0.9615 & 0.9625 & 0.972 \\ \hline
$L=30$  & 0.958 & 0.9615 & 0.9715 \\ \hline
    \end{tabular}
    \begin{tabular}{ | l | l | l | l |}
    \hline
    100 training samples  &  MLP NTK & ResNet NTK $\alpha=1$ & ResNet NTK $\alpha=0.1$ \\ \hline
  $L=5$  & 0.981 & 0.9885 & 0.9905 \\ \hline
$L=15$  & 0.973 & 0.982 & 0.9925 \\ \hline
$L=30$  & 0.9645 & 0.9795 & 0.991 \\ \hline
    \end{tabular}}
\end{table}

\newpage

\section{Closed-Form $T$ and $\dot{T}$ Expressions for ReLU Nonlinearities}
\label{app:T_for_relu}

For completeness, we present here the closed-form expression of $T(\K)$ and $\dot{T}(\K)$ for $\phi(\cdot)$ which is the ReLU activation function. These results are due to \citep{cho2009kernel}.

Let $\K := \left  [ \begin{matrix} K_{11} & K_{12} \\
  K_{12} & K_{22} \\
\end{matrix} \right  ]$ be a $2 \times 2$ positive semidefinite matrix, $\rho:=\frac{K_{12}}{\sqrt{K_{11} K_{22}}}$, and recall the definitions
$T(\K):=\Exppp{ ( u,v )  \sim \mathcal{N}(\0,\K)}{\phi(u) \phi(v)}$
and 
$\dot{T}(\K):=\Exppp{ ( u,v )  \sim \mathcal{N}(\0,\K)}{\phi'(u) \phi'(v)}$.
For the special case where $\phi(\cdot)=\mathrm{max}\{0,\cdot\}$, we have
\begin{align}
\label{Eq_T_for_relu}   
T(\K) &=\frac{ 1 }{2\pi} \sqrt{K_{11} K_{22}} \left ( \rho \left ( \pi - \mathrm{arccos} \left ( \rho \right ) \right ) + \sqrt{1-\rho^2} \right ), \\ \nonumber
\dot{T}(\K) &=\frac{1}{2\pi} \left ( \pi - \mathrm{arccos} \left ( \rho \right ) \right ).
\end{align}

\begin{figure}
  \centering
  \subfigure[{\scriptsize NTKs (normalized to unit peak) $L=5$}]{%
    \label{fig:erf_NTK_mlp_resnet_L5_kernels}
    \includegraphics[width=140pt]{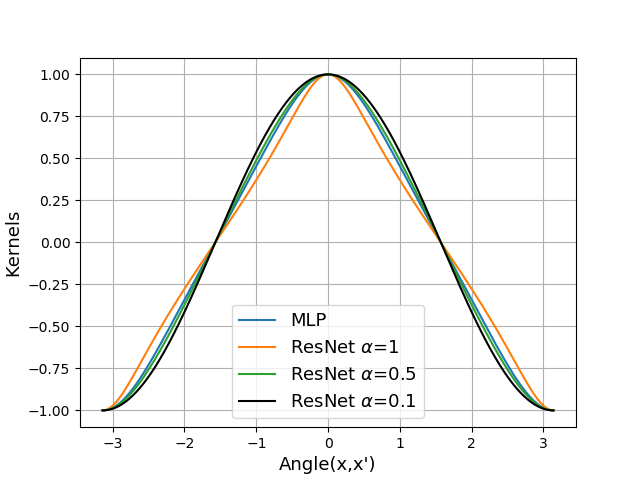}
  }
  \subfigure[{\scriptsize Interpolation with 6 samples $L=5$}]{%
    \label{fig:erf_NTK_mlp_resnet_L5_regression_N6}
    \includegraphics[width=140pt]{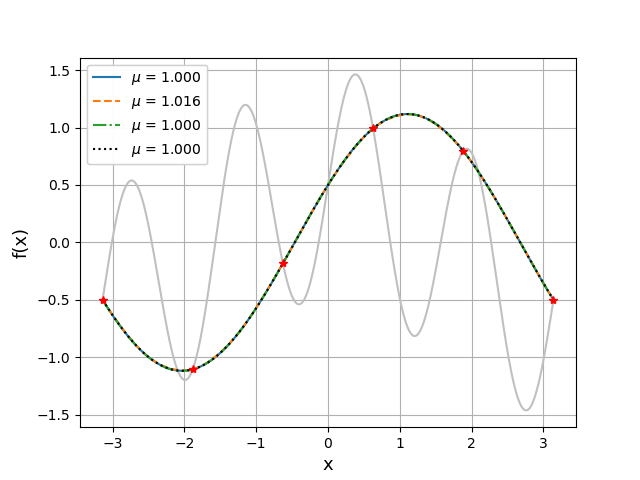}
  }
  \subfigure[{\scriptsize Interpolation with 10 samples $L=5$}]{%
    \label{fig:erf_NTK_mlp_resnet_L5_regression_N10}
    \includegraphics[width=140pt]{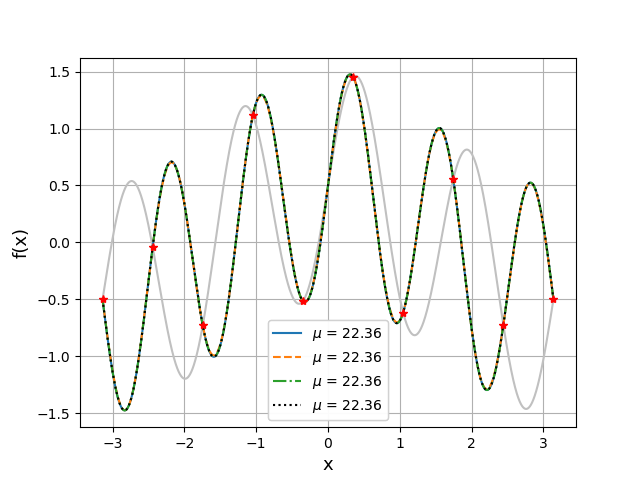}
  }

    \caption{NTKs for MLP and ResNet (for different values of $\alpha$) with $L=5$ nonlinear layers, erf nonlinearities, $\sigma_v=\sigma_w=1$, for inputs on the sphere (circle) in $\mathbb{R}^2$.
    (\ref{fig:erf_NTK_mlp_resnet_L5_kernels}): The kernels shape.  
    (\ref{fig:erf_NTK_mlp_resnet_L5_regression_N6})-(\ref{fig:erf_NTK_mlp_resnet_L5_regression_N10}): Interpolations by the closed-form solutions, measured by $\mu(\cdot)$ defined in \eqref{Eq_L2_der2_meas2}. 
    Note that the legend in (\ref{fig:erf_NTK_mlp_resnet_L5_kernels}) applies to all the figures. 
    }
\label{fig:NTK_mlp_resnet_L5_erf}     
\end{figure}

\section{\tomtb{NTK Experiments with ERF Nonlinearities}}
\label{app:T_for_erf}

\tomtb{
In this section, we present NTK experiments for the erf activation function.
First, we present the closed-form expression of $T(\K)$ and $\dot{T}(\K)$ for $\phi(\cdot)$ which is the erf function.} 

\tomtb{
Let $\K := \left  [ \begin{matrix} K_{11} & K_{12} \\
  K_{12} & K_{22} \\
\end{matrix} \right  ]$ be a $2 \times 2$ positive semidefinite matrix and recall the definitions
$T(\K):=\Exppp{ ( u,v )  \sim \mathcal{N}(\0,\K)}{\phi(u) \phi(v)}$
and 
$\dot{T}(\K):=\Exppp{ ( u,v )  \sim \mathcal{N}(\0,\K)}{\phi'(u) \phi'(v)}$.
For the special case where $\phi(u)= \frac{2}{\sqrt{\pi}} \int_{0}^{u} \mathrm{e}^{-z^2}dz$, we have from \citep{williams1996computing} that 
\begin{align}
\label{Eq_T_for_erf}   
T(\K) &=\frac{ 2 }{\pi} \mathrm{arcsin \left ( \frac{2 K_{12}}{ \sqrt{(1+2 K_{11})(1 + 2 K_{22})} }  \right )}, \\ \nonumber
\dot{T}(\K) &=\frac{4}{\pi} \mathrm{det} \left ( \I_2 + 2 \K \right )^{-1/2}.
\end{align}}

\tomtb{
Next, we repeat several NTK experiments from Section~\ref{sec:ntk_resnet_vs_mlp__regression},
but with the erf activation instead of the ReLU activation. The other configurations are not changed.
The results are presented in Figure~\ref{fig:NTK_mlp_resnet_L5_erf}.
It can be seen that with erf activations both MLP and ResNet NTKs have rather similar shapes and similar interpolation results (contrary to the ReLU case). This may imply that the smoothness distinction between the models requires using nonsmooth activations.}

\end{document}